\documentclass[lettersize,journal]{IEEEtran}
\usepackage{amsmath,amsfonts}
\usepackage{algorithmic}
\usepackage{algorithm}
\usepackage{array}
\usepackage[caption=false,font=normalsize,labelfont=sf,textfont=sf]{subfig}
\usepackage{textcomp}
\usepackage{stfloats}
\usepackage{url}
\usepackage{verbatim}
\usepackage{graphicx}
\usepackage{cite}
\usepackage{booktabs}
\usepackage{threeparttable} 
\usepackage{upgreek}
\usepackage{color}
\usepackage{multirow}

\usepackage{multirow}

\makeatletter
\let\NAT@parse\undefined
\makeatother
\usepackage[colorlinks,linkcolor=red,anchorcolor=blue,urlcolor=magenta,citecolor=blue]{hyperref}

\begin{document}

\title{Review on Panoramic Imaging and Its Applications in Scene Understanding}

\author{Shaohua Gao$^{1}$, Kailun Yang$^{2}$, Hao Shi$^{1}$, Kaiwei Wang$^{1}$, and Jian Bai$^{1}$ 
\thanks{This work was supported in part by the National Natural Science Foundation of China (NSFC) under Grant No. 12174341, in part by the Federal Ministry of Labor and Social Affairs (BMAS) through the AccessibleMaps project under Grant 01KM151112, in part by the University of Excellence through the ``KIT Future Fields'' project, in part by the Hangzhou Huanjun Technology Company Ltd., and in part by Hangzhou SurImage Technology Company Ltd.
\emph{(Corresponding authors: Kaiwei Wang and Kailun Yang.)}}
\thanks{$^{1}$S. Gao, H. Shi, K. Wang, and J. Bai are with State Key Laboratory of Modern Optical Instrumentation, Zhejiang University, China {\tt\small \{gaoshaohua, haoshi, wangkaiwei, bai\}@zju.edu.cn}}
\thanks{$^{2}$K. Yang is with Institute for Anthropomatics and Robotics, Karlsruhe Institute of Technology, Germany {\tt\small kailun.yang@kit.edu}}%
\thanks{Manuscript received May 11, 2022.}}

\markboth{IEEE Transactions on Instrumentation and Measurement, October~2022}%
{Gao \MakeLowercase{\textit{et al.}}: Review on Panoramic Imaging and Its Applications in Scene Understanding}

\maketitle

\begin{abstract}
With the rapid development of high-speed communication and artificial intelligence technologies, human perception of real-world scenes is no longer limited to the use of small Field of View (FoV) and low-dimensional scene detection devices. Panoramic imaging emerges as the next generation of innovative intelligent instruments for environmental perception and measurement. However, while satisfying the need for large-FoV photographic imaging, panoramic imaging instruments are expected to have high resolution, no blind area, miniaturization, and multidimensional intelligent perception, and can be combined with artificial intelligence methods towards the next generation of intelligent instruments, enabling deeper understanding and more holistic perception of $360^\circ$ real-world surrounding environments. Fortunately, recent advances in freeform surfaces, thin-plate optics, and metasurfaces provide innovative approaches to address human perception of the environment, offering promising ideas beyond conventional optical imaging. In this review, we begin with introducing the basic principles of panoramic imaging systems, and then describe the architectures, features, and functions of various panoramic imaging systems. Afterwards, we discuss in detail the broad application prospects and great design potential of freeform surfaces, thin-plate optics, and metasurfaces in panoramic imaging. We then provide a detailed analysis on how these techniques can help enhance the performance of panoramic imaging systems. We further offer a detailed analysis of applications of panoramic imaging in scene understanding for autonomous driving and robotics, spanning panoramic semantic image segmentation, panoramic depth estimation, panoramic visual localization, and so on. Finally, we cast a perspective on future potential and research directions for panoramic imaging instruments.

\end{abstract}

\section{Introduction}
\IEEEPARstart{R}{ecent} advances in high-speed communication and artificial intelligence have led to stronger demand for upgrading traditional imaging systems. Panoramic imaging has a wider Field of View (FoV) than traditional panoramic optical systems and can capture more information about the surrounding environment at a time. It is becoming the next generation of intelligent sensing instruments. They are used in autonomous driving, machine vision inspection, endoscopic medicine, satellite atmospheric inspection, and many other applications. In terms of imaging performance, panoramic imaging faces several common challenges to satisfy the requirements of applications such as machine vision, including FoV, resolution, no blind area, and imaging quality. Another urgent requirement is multidimensional intelligent perception, where panoramic imaging is expected to be combined with intelligent sensors to record, fuse, and perceive higher dimensional information about the surrounding environment. In addition, panoramic imaging is expected to evolve towards lightweight and small-volume structural forms for applications in more space- and weight-constrained scenarios.
However, the above requirements usually entail addressing different trade-offs, which make the design of panoramic imaging instruments particularly challenging.
Compared with the only previous panoramic imaging survey~\cite{Gledhill2003PanoramicI}, it is so far back that it does not provide an overview of the flourishing progress in panoramic imaging in the last 20 years~\cite{bonarini2000omnidirectional_tracking,jiang2021500FPS,liu2021blind_omnidirectional,kholodilin2020omnidirectional_laser,cao2021quality_measurement,hu2022distortion}, which is addressed in this review.
Surveys on other sensing have emerged, such as LiDAR~\cite{roriz2021automotive_lidar,gao2021we_hungry_lidar} or applications in scene understanding~\cite{feng2021deep_multimodal,zhang2021deep_multimodal_fusion_survey,ming2021deep_monocular_depth_estimation_review}. However, a detailed review of panoramic imaging and its applications has not appeared so far. This review fills the gap by combing panoramic imaging and its scenario applications from their origin to the latest developments.
Panoramic imaging experienced its first boom in the 1990s, but this boom gradually declined due to the lack of qualified manufacturing equipment and digital information processing systems.

\begin{figure*}[!t]
\centering
\subfloat[]{\includegraphics[width=6in]{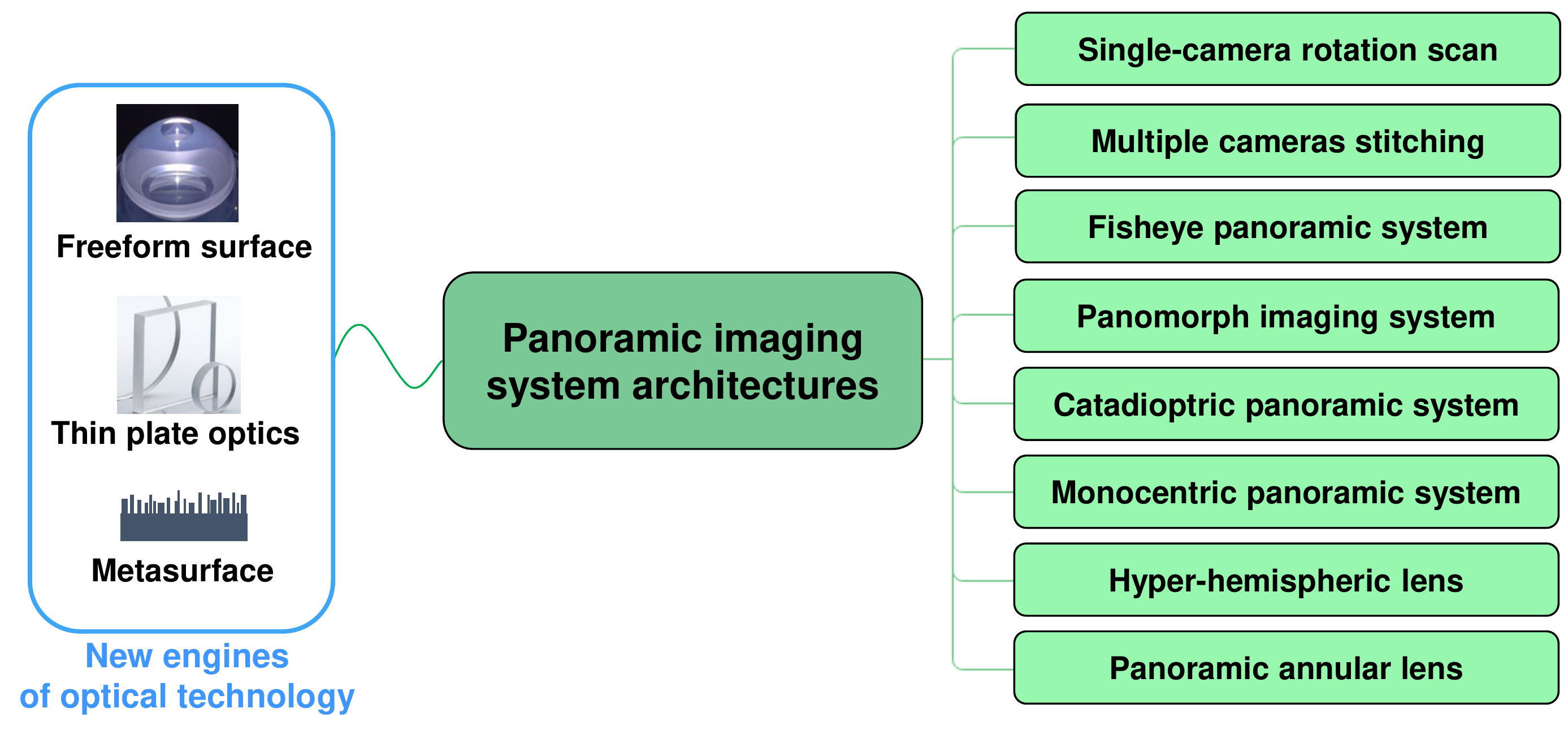}}%
\hfil
\subfloat[]{\includegraphics[width=3in]{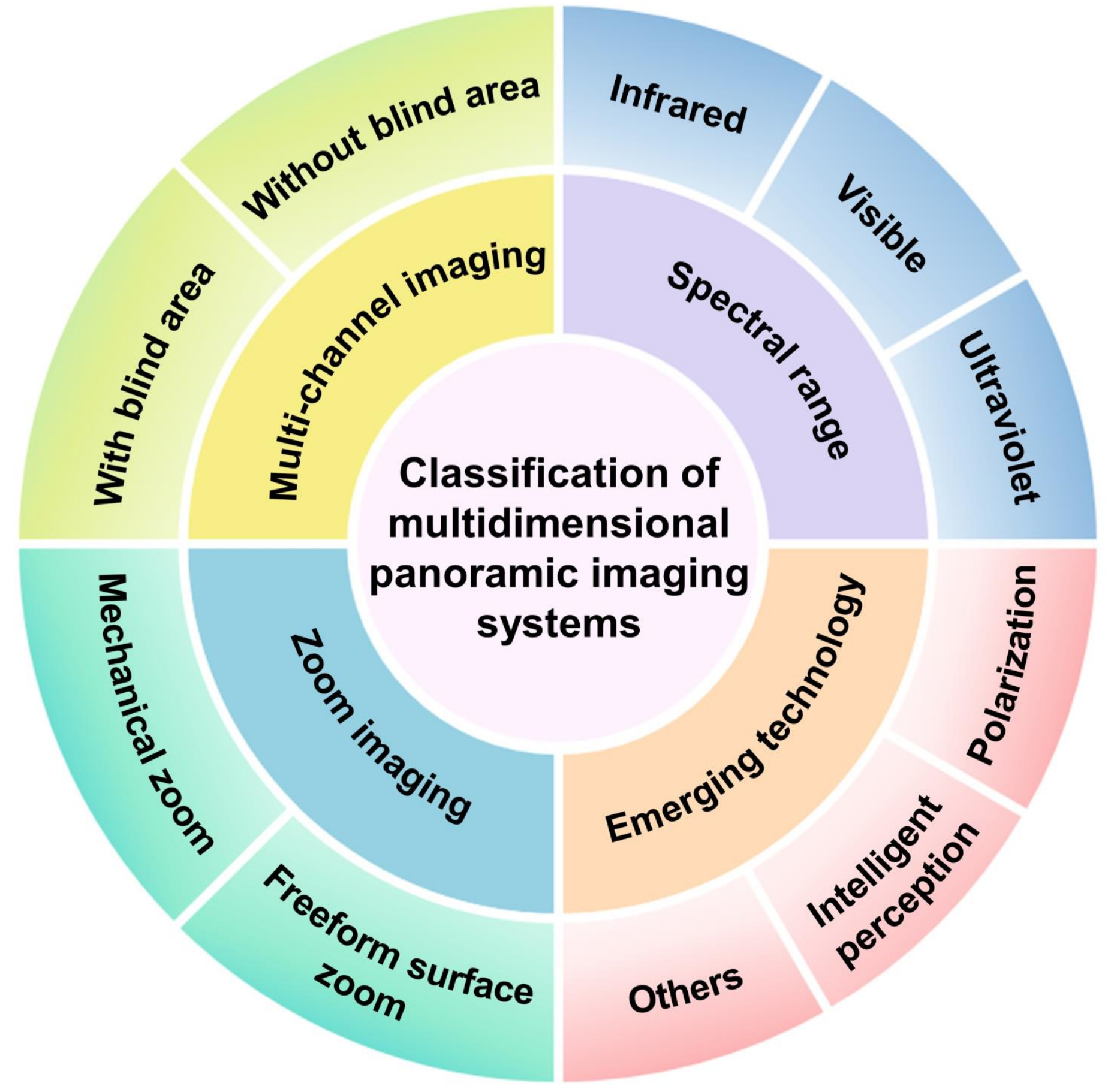}}%
\caption{Schematic of panoramic imaging systems. (a) Panoramic imaging system architectures; (b) Classification of multidimensional panoramic imaging systems.}
\label{fig_1}
\end{figure*}

In the past decade, the concept of panoramic imaging has been revisited and has enjoyed a new boom. Emerging technologies such as freeform surfaces~\cite{Rolland2021FreeformOF} and metasurfaces~\cite{Shalaginov2020SingleElementDF} have dramatically reshaped panoramic imaging systems and lit a bright light in the panoramic imaging field.
The application of these emerging optical technologies has made a strong impetus for the performance improvement of panoramic optical system architectures (Fig.~\ref{fig_1}(a)). At the same time, the proposal of the multidimensional panoramic imaging system has enriched the panoramic imaging and played a more powerful role in different application fields (Fig.~\ref{fig_1}(b)). In this review, we first introduce the basic concepts of panoramic imaging, related design challenges, and potential solutions. Afterwards, we review different architectures of panoramic imaging systems and their characteristics in detail. In addition, we classify and analyze different panoramic imaging systems according to their multidimensional imaging modalities and briefly introduce the emerging optical technologies that are currently being equipped for panoramic imaging. Then, we summarize and discuss the applications of panoramic imaging in scene understanding in detail. Finally, this review provides a comprehensive overview and outlook in the field of panoramic imaging.

\section{Key Parameters of Panoramic Imaging Systems}
To satisfy the panoramic environment perception, panoramic imaging systems usually face several common challenges, such as field of view, wavelength, F-number, focal length, resolution, imaging quality, and volume~\cite{Wang2019DesignOH,Gao:22}. To achieve panoramic imaging systems that meet application scenarios, these key parameters and their trade-offs usually need to be considered before design. Potential solutions can facilitate the design and manufacture of panoramic imaging systems with higher parametric performance. Before discussing each parameter in detail, it is helpful to review the basic definitions of the above parameters.
\subsection{Definition of Parameters}
For panoramic imaging systems, the performance parameters of the optical system are particularly important. In this section, the definitions of the key parameters of panoramic imaging systems are summarized.

\begin{figure}[!t]
\centering
\includegraphics[width=3in]{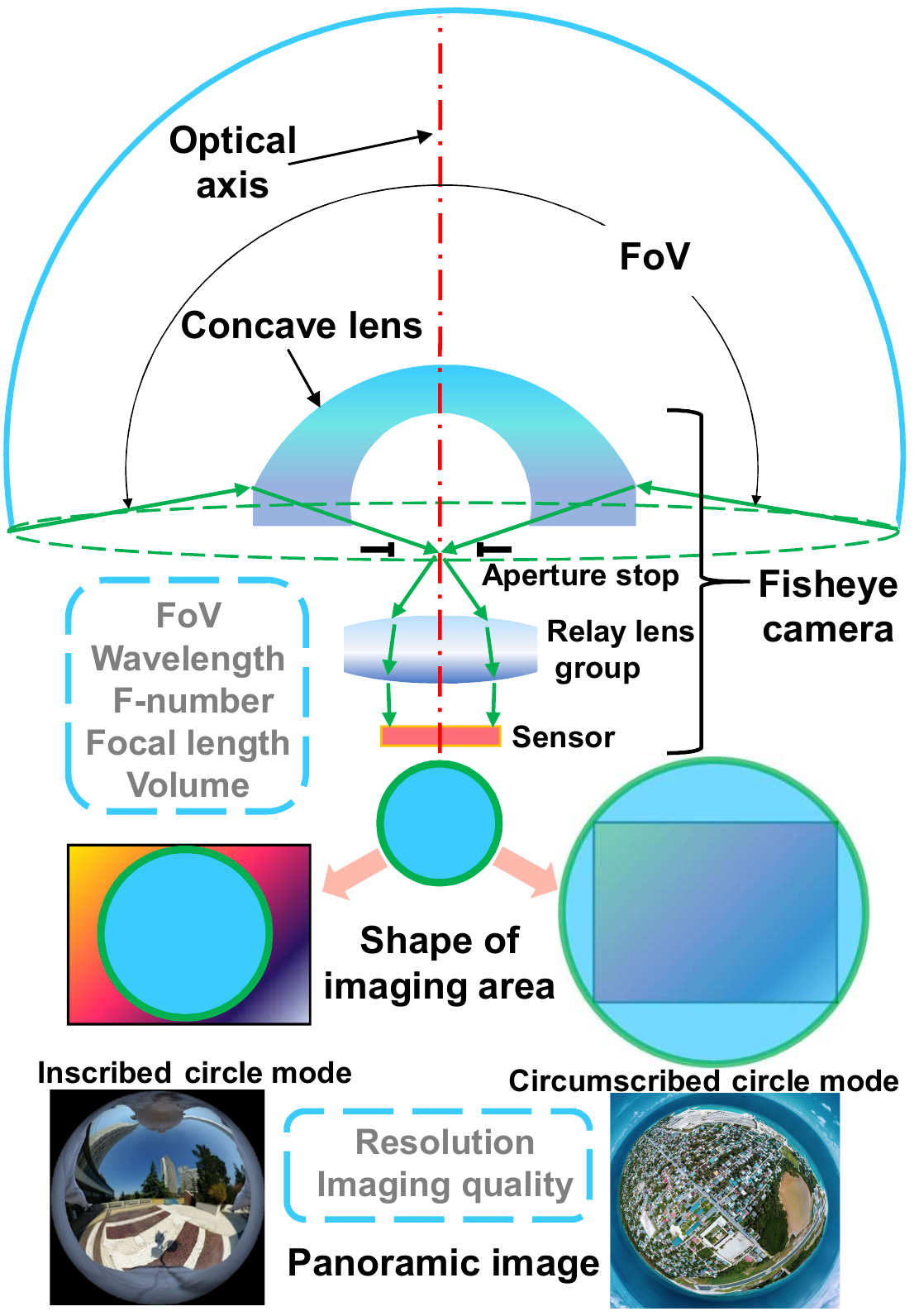}
\caption{Parameter trade-off of panoramic optical system (Take a fisheye camera as an example). Panoramic imaging modes: Inscribed circle mode (bottom left corner) and Circumscribed circle mode (bottom right corner). Reproduced with permission~\cite{Hwang2019TowardHM}. Copyright 2019, IEEE.}
\label{2}
\end{figure}

1) Field of View (FoV): In applied optics, there are four ways to express the FoV: angle, object height, paraxial image height, and actual image height. Panoramic imaging instruments usually take the angle as the evaluation standard of the FoV. In practical applications, full angle FoV is generally used. When designing a panoramic optical system, the expression of $360^\circ{\times}$(Half FoV) is also usually used to represent a panoramic system with a blind area. In optical instruments, the angle formed by the two edges of the maximum range through which the object of the measured target can pass the camera, with the camera of the optical instrument as the vertex, is called the Field of View (FoV) \cite{geary2002introduction,Song2019Introduction}. 
Here, we take the most common fisheye optical system (Fig.~\ref{2}) as an example. The FoV determines the ability of the optical instrument to observe the range of the surrounding scene.
Compared to other common imaging optical systems, the FoV of a fisheye optical system can usually reach $180^\circ$ or more.
This unique imaging characteristic enables more information about the surrounding environment to be recorded at a time.
Common imaging systems typically have a small FoV that captures less information each time.
Therefore, panoramic imaging systems have more environmental perception advantages over traditional optical systems with common FoVs and have become a popular development direction for future optical instruments.
The concept of metaverse has brought Augmented Reality (AR) and Virtual Reality (VR) display fields into a new round of development, and has also led to the rise of panoramic environment perception~\cite{Xiong2021AugmentedRA,Dewen2021DesignAM}.
New panoramic imaging instruments enables a broader environmental perception of the real world and transmits the data to artificial intelligence and big data processing for further human awareness and understanding of real-world scenes.

2) Wavelength: In physics, wavelength is the spatial period of a periodic wave~\cite{Hecht1998Optics4E}. Electromagnetic radiation can be classified by wavelength as radio waves, microwaves, infrared, visible spectrum, ultraviolet, X-rays, and Gamma rays~\cite{Sliney2016WhatIL,Arnold2020TheES}. The spectral range in which the human eye can perceive the world can be described as visible light, usually defined as wavelengths in the range of $400{\sim}700$ nanometers (nm)~\cite{Arnold2020TheES}. Light wavelengths longer than the visible spectrum are called infrared. Light wavelengths shorter than the visible spectrum are called ultraviolet. Ultraviolet (UV) wavelengths range from $10$ nm to $400$ nm and are shorter than visible light, but longer than X-rays~\cite{Tobiska2006ISO2}. Typically, infrared (IR) wavelengths range from $700$ nm to $1$ millimeter (mm)~\cite{Cyan1970HandbookOC}. Visible light wavelengths are the most common range for designing panoramic imaging optical systems~\cite{Gao:22}. To expand human perception of the surrounding environment, UV and IR panoramic imaging systems have emerged to enable a wider spectrum of environmental perception beyond the human visible spectrum wavelengths~\cite{Liping2011DesignOC,Powell1996DesignSO}.

3) F-number: It is defined as the ratio of the focal length (in the image space) of an optical system to its entrance pupil diameter~\cite{greivenkamp2004field,bass2010handbook}.
The smaller the F-number, the more light is fed into the optical system and the larger the size of the aperture diaphragm (aperture stop). Small F-number optical systems are also called high-speed imaging systems. Small F-number optical systems generally have a large aperture and greater luminous flux, which can improve shutter speed. In low light condition, this optical system can maintain more light flux, which is good for a night- or darkfield shooting~\cite{Holst2012SmallDI}.
When shooting moving objects, a small F-number optical system is more suitable for shooting the object clearly with a high-speed shutter.
Optical system with small F-number has small depth of field~\cite{ROBINSON2016201}. When shooting, the optical system with smaller depth of field will highlight the target while defocusing the background. A large F-number optical system has a large depth of field, enabling clear imaging of a wide range of scenes at both distant and close range, but with reduced illumination.
F-number is a crucial parameter. The smaller the F-number, the wider the scope of applications of the camera. When designing an imaging system, a small F-number optical system is more difficult to design. But it typically has higher imaging quality.
Therefore, the F-number of a panoramic system is usually designed to be small enough
to achieve a large luminous flux and excellent imaging quality.

4) Focal length: The focal length is the distance from the primary point of the optical system to the focal point\cite{goodman1996general}. The review of the projection models can help us understand the principles of panoramic imaging.
Optical design usually follows five classical projection models~\cite{bettonvil2005fisheye,Schneider2009ValidationOG}.
They are rectilinear projection model, equidistant projection model, equisolid angle projection model, orthographic projection model, and stereographic projection model.
The equidistant projection model is widely used in the design of large FoV optical systems, and its expression in Equation (\ref{deqn_ex1}) is:
\begin{equation}
\label{deqn_ex1}
y = f \cdot \theta,
\end{equation}
where $f$ is the focal length, $\theta$ is the half FoV, and $y$ is the image height. The selection of the sensor is required to be matched with the optical system. Similar to other panoramic optical systems, the fisheye optical system contains two imaging modes: the Inscribed circle mode is illustrated in the bottom left corner of Fig.~\ref{2} and the Circumscribed circle mode is illustrated in the bottom right corner of Fig.~\ref{2}. The rectangle is the sensor and the circle is the imaging circle of the fisheye optical system in the image plane. Panoramic images of these two imaging modes are presented below the corresponding imaging mode schematic. A panoramic system designed with the Inscribed circle imaging mode do not lose the FoV, but waste pixels at the edge of the sensor. To achieve panoramic scene understanding with the largest FoV, this mode is used more frequently in the design stage of panoramic systems. Panoramic systems designed with Circumscribed circle imaging modes lose the FoV at the edges. Sometimes this solution is adopted in the design of optical systems without central blind areas such as fisheye optical systems. When the FoV of a panoramic optical system remains the same value, the larger the focal length and the larger the image height. At the same time, the volume of the designed system will be larger. Its matched sensor tends to have a higher resolution (In the case of a constant sensor pixel size).

5) Resolution: Image resolution generally refers to the ability of a measurement or display system to resolve details. This concept can be measured in the fields of time, space, \textit{etc.} In panoramic system design, resolution generally refers to the total number of arrays of horizontal and vertical pixels of the sensor, such as $4216(H){\times}3128(V)$ for a $13$ megapixel sensor~\cite{warren2008modern,Dudzik1993TheI,Carlson2002ComparisonOM,Lenz1994NewDI}. The pixel size and resolution of the sensor jointly determine the optical format of the sensor~\cite{bockaert2002sensor,Benjamin2016PracticalOT}. For the optical system, optical resolution is relevant to the wavelength and aperture diaphragm size. In specific, the shorter the wavelength and the larger the aperture diaphragm, the higher the resolution of the optical system (In the case of great system aberration correction).

6) Imaging quality: The imaging quality of a panoramic optical system is not a single evaluation criterion. The biggest influence on imaging quality is aberration. Aberrations are mainly divided into spherical aberration, coma, astigmatism, field curvature, distortion, chromatic aberration, and wave aberration~\cite{greivenkamp2004field}. Chromatic aberration can be divided into longitudinal chromatic aberration and lateral chromatic aberration. Longitudinal chromatic aberration is also known as positional chromatic aberration, and lateral chromatic aberration is also known as transverse chromatic aberration. In the process of designing a panoramic optical system, the image quality of a panoramic imaging system is usually evaluated by the spot diagram and the Modulation Transfer Function (MTF)~\cite{Gao:22}. The optical system needs to follow the aberration design theory to correct the aberration in the design process, and the final design needs to satisfy the uniform relative illumination of each FoVs~\cite{Uvarova2021OpticalDO,Tsyganok2021DevelopmentOM}. In order to make the panoramic imaging instrument more conducive to scene understanding, the relative illumination of the panoramic optical system also needs to be as uniform as possible, and the distortion should be as small as possible to meet the perception accuracy requirements~\cite{Gao:22}.

7) Volume: Miniaturized and lightweight panoramic optical systems will be suitable for more space and weight-constrained scenarios~\cite{Gao:22}. To facilitate the characterization and comparison of the compactness of panoramic systems, a better way is to use the panoramic system compactness ratio as the volume compactness parameter.

The compact ratio of the panoramic system as shown in Equation (\ref{deqn_ex2}) can be defined as the ratio of the maximum diameter of the panoramic system to the diameter size of an image on the sensor~\cite{Luo2017CompactPD}:
\begin{equation}
\label{deqn_ex2}
R_{compact} =\frac{D_{panoramic}} {D_{sensor}},
\end{equation}
where $R_{compact}$ is the ratio of compactness, $D_{sensor}$ is the maximum diameter of the imaging circle on the sensor, and $D_{panoramic}$ is the maximum lateral diameter of the panoramic optical system.

At present, researchers in the panoramic field are inclined to design small volume, small F-number, and high-resolution panoramic optical systems to satisfy the demand for higher optical performance and compactness for panoramic field photography. At the beginning of the design, the parameters need to be considered and weighed in detail.

\begin{table*}[!t]  
\caption{Comparison of different panoramic imaging systems.}  
	\label{tab:panoramic imaging systems_comparison} 
	\centering
	\resizebox{0.95\textwidth}{!}{
    \renewcommand\arraystretch{1.6}{\setlength{\tabcolsep}{0.5mm}{\begin{tabular}{c|c|c|c|c|c}  
	
	\hline
		System category&Dynamic scene&Image stitching&Compactness&Blind area&Distortion (In most cases)\\ 

	\hline
	Single-camera rotation scanning~\cite{Gledhill2003PanoramicI,Wagner2010RealtimePM,Lin2014FrontVA}&No&Yes&Low&No&Low\cr
	
	\hline
	
	Multiple cameras stitching~\cite{Yuan2010ANM,Cowan2019360SI,Lin2019ALP}&Yes&Yes&Low&No&Low\cr
	
	\hline
	
	Fisheye panoramic system~\cite{Thibault2005EnhancedOD,ning2014wide,Fan:19}&Yes&No&Medium&No&15\%$\sim$20\% (Equidistant projection)\cr
	
	\hline
	
	Panomorph imaging system~\cite{Thibault2006PanomorphLC,Thibault2006PanomorphLA,Thibault2014DesignFA}&Yes&No&High&No&Distortion correction\cr
	
	\hline
	
	Catadioptric panoramic system~\cite{Zhuang2019,2019Design}&Yes&No&Medium&Yes&$\sim$10\% or more (Equidistant projection)\cr
	
	\hline
	
	\multirow{2}{*}{Monocentric panoramic system~\cite{Huang2020ModelingAA,Tremblay2012DesignAS,stamenov2014broad,wang2019design}}&\multirow{2}{*}{Yes}&\multirow{2}{*}{Yes}&\multirow{2}{*}{High}&\multirow{2}{*}{No}&1\%$\sim$3\% (Monocentric multi-scale imager, Rectilinear projection)\cr
	
	&&&&&$\sim$20\% (Monocentric lens imager, Equidistant projection)\cr
	
	\hline
	
	Hyper-hemispheric lens~\cite{pernechele2018introduction,Pernechele2013HyperhemisphericAB,Pernechele2016HyperHL}&Yes&No&High&Yes$/$No&Distortion correction\cr 
	
	\hline
	
	Panoramic annular lens~\cite{Zhou2020DesignAI,gao2021design,wang2022high_performance,Gao:22}& Yes&No&High&Yes$/$No&1\%$\sim$5\% (Equidistant projection)\cr  
	\hline
	\end{tabular}}}}
 
\end{table*}

\subsection{Key Parameters Trade-offs and Potential Solutions}
In the design process of panoramic optical systems, a larger FoV is usually pursued~\cite{Zhang2020DesignOA}. The increased FoV will bring greater difficulty in aberration correction.
The influence of the aberration of the panoramic imaging system on the imaging quality will be solved in the hardware design stage. Therefore, it is particularly important to evaluate which panoramic system architecture to be used in the optical design process to obtain panoramic images with high imaging quality and a large FoV.
At the beginning of the development of panoramic optical systems, multiple cameras stitching~\cite{Yuan2010ANM,Cowan2019360SI} or single camera scanning~\cite{Gledhill2003PanoramicI} techniques were usually used. Single-camera scanning requires high speed, high frame rate, and high precision to maintain scanning stability.
However, this method cannot capture the surrounding environment in real time~\cite{Brown2007MinimalSF}.
On the other hand, multiple cameras stitching technology needs to rely on post-algorithm stitching and correction, and there are still errors in camera calibration and correction~\cite{Yuan2010ANM}. The multiple cameras stitching approach usually brings greater device volume and power consumption, as well as the inevitable high price and arithmetic consumption. Subsequent developments in the fisheye optical system use a refractive optical path for single-camera large FoV imaging~\cite{ning2014wide,Kim2014ToleranceAA}. The optical path to satisfy the large FoV aberration correction makes the optical path smooth, usually using more negative lenses in the front to deflect the direction of the light path for large FoV~\cite{Pernechele2021TelecentricFF}. To reduce the number of lenses used, the difficulties of design, processing, and assembly, high refractive index materials are usually used in the design~\cite{Fan:19}. Although highly refractive materials have a greater ability to refract light, they tend to be more expensive. To further reduce the volume of the system, new panoramic imaging architectures such as catadioptric panoramic system~\cite{2019Design}, hyper-hemispheric lens~\cite{Pernechele2013HyperhemisphericAB}, panoramic annular lens~\cite{Gao:22}, \textit{etc.} have emerged. These new architectures usually use a catadioptric optical path to collect the light from a large FoV into a relay lens group and then perform aberration correction. The parameter comparison of different panoramic imaging system architectures is shown in Table~\ref{tab:panoramic imaging systems_comparison}.
Each of these panoramic imaging systems has its own architecture, which will be discussed in detail in the next section.

To achieve higher relative illumination and higher optical resolution, small F-number panoramic optical systems have emerged~\cite{gao2021design}. Long focal length panoramic imaging systems~\cite{Niu2007DesignOA,Wang2019DesignOH} are able to match high-resolution imaging sensors, thus, enabling detailed observation of the panoramic environment. The human eye can only perceive a very narrow spectrum of visible light in electromagnetic waves, which is approximately $400{\sim}700$ nm~\cite{Arnold2020TheES}. To provide higher-dimensional detection of the real environment, panoramic imaging systems in the infrared~\cite{Yao2016DesignOA,Cowan2019360SI,2019Design}, visible~\cite{Wang2015DesignOP,Zhang2020DesignOA}, and ultraviolet~\cite{Liping2011DesignOC,qingsheng2014optical} wavelengths have been designed in succession. Not only from the light wave spectrum but also the polarized light~\cite{Luo2017CompactPD,gao2021design} can be used for the design of panoramic systems. The use of panoramic optical systems to broaden the human perception of the surrounding environment has become a new trend in the development of panoramic imaging systems. Multidimensional panoramic perception will broaden human understanding of the real environment and provide a deeper understanding and awareness of the real world. Meanwhile, zoom panoramic optical systems~\cite{Ma2011DesignOA,Wang2021DesignOA} can switch between large and small FoVs in the panoramic range to achieve detailed detection of key areas of interest. To obtain a wider observation FoV, multi-channel panoramic systems~\cite{Huang2017DesignOA,Liu2019DesignOC} have also been proposed. These multi-channel panoramic imaging systems usually have larger FoVs with different views. They have certain advantages over single-channel panoramic systems. Catadioptric panoramic optical system usually have a central blind area~\cite{pernechele2018introduction}. To eliminate this tricky drawback, dichroic films~\cite{Luo2016NonblindAP} and polarization techniques~\cite{Luo2017CompactPD,gao2021design} were introduced to eliminate the blind areas. Conventional panoramic systems are designed using spherical surfaces~\cite{Wang2019DesignOH,wang2022high_performance}. The optical lens design techniques use the accumulation of optical paths to modulate the beam~\cite{Yu2014FlatOW}. As the FoV increases, it becomes more difficult to correct the optical aberration and more lenses are usually used. Such techniques make the design and manufacture of higher-performance compact panoramic systems more difficult. It also makes assembly more difficult and brings an inevitable increase in volume and price.

Fortunately, the rapid development of high-precision manufacturing technology has made new types of optical surfaces possible. Freeform surface~\cite{Ma2011DesignOA,YangTong2021FreeformIO}, thin-plate optics~\cite{Peng2019LearnedLF}, and metasurface~\cite{Shalaginov2019ASP} technologies provide powerful engines for miniaturization and high imaging performance of panoramic imaging systems.
These technologies offer more freedom degrees to optimize the design of panoramic optical systems, allowing for better image quality to an extreme extent, while reducing the system volume~\cite{YangTong2021FreeformIO}.
In application areas, panoramic imaging devices are becoming promising filming devices for autonomous driving, medical treatment, satellite navigation, \textit{etc}~\cite{gao2021design,wang2022pal_slam}. Combined with artificial intelligence and multidimensional sensor technology, panoramic imaging will have broader application scenarios and become a new trend in photographic instruments.

\begin{figure*}[!t]
\centering
\subfloat{\includegraphics[width=7in]{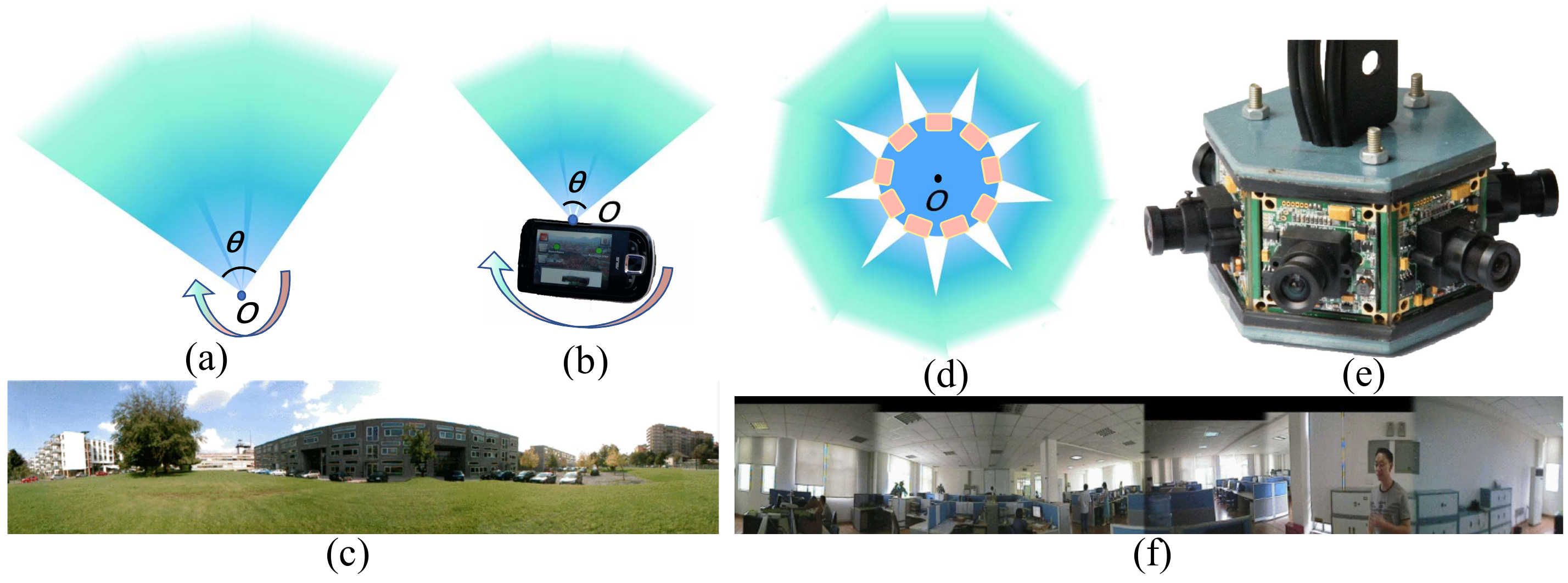}}%
\hfil
\caption{Panoramic images using FoVs stitching technology. (a) Principle of single-camera scanning shooting; (b) A method of using a mobile phone to capture large-FoV images with a single camera; (c) A large-FoV image generated by a mobile phone. (b) and (c) are reproduced with permission~\cite{Wagner2010RealtimePM}. Copyright 2010, IEEE; (d) Principle of multiple cameras stitching; (e) A case of FoVs stitching using six visible light cameras; (f) Imaging result of six circumferential arrays visible light cameras. (e) and (f) are reproduced with permission~\cite{Yuan2010ANM}. Copyright 2010, IEEE.}
\label{fig_3}
\end{figure*}

\section{Architectures and Properties of Panoramic Imaging Systems}
\subsection{Single Camera Scanning and Multiple Cameras Stitching}
FoVs stitching is a method of using multiple FoVs to capture and stitch together to achieve large-FoV images~\cite{Gurrieri2013AcquisitionOO,Ji2020PanoramicSF,Scaramuzza2021OmnidirectionalC}.
This technology is implemented in single-camera scanning and multi-camera stitching methods. Single-camera scanning uses a high-precision mechanical rotating stage to rotate a single camera to create a panoramic view. The shooting principle is shown in Fig.~\ref{fig_3}(a), where the optical system is scanned around the center point $O$ to form an image with a large FoV.
Using a mobile phone, a large-FoV panoramic image can be captured, as shown in Fig.~\ref{fig_3}(b), and the captured image is illustrated in Fig.~\ref{fig_3}(c)~\cite{Wagner2010RealtimePM}.
A similar single-camera scan case is the installation of Nikon 990 mounted on Kaidan Kiwi 990.
It can form a large FoV image of $120^\circ$\cite{Gledhill2003PanoramicI}. This imaging method can be applied to shoot scenes in a relatively static environment, which requires high scanning accuracy for mechanical devices. The camera cannot achieve gaze imaging because it takes some time to rotate and scan during the shooting.
Using a prism to flip the FoV and rotate the shot is a similar scheme~\cite{Lin2014FrontVA}.

Another common FoV-stitching technique is to use multiple cameras for stitching shots~\cite{Huang2014A3P}.
Multiple cameras shot simultaneously to achieve $360^\circ$ surround-view image, and the shooting schematic is shown in Fig.~\ref{fig_3}(d).
In~\cite{Yuan2010ANM}, a case of FoVs stitching using six visible light cameras is discussed (Fig.~\ref{fig_3}(e)). The captured image is shown in Fig.~\ref{fig_3}(f). Due to the inconsistency of the parameter settings and installation angles of each camera, the image stitching is affected. Long-wave infrared optical materials are usually costly.
Adopting the multi-camera stitching panoramic imaging technique, Cowan~\textit{et al.}~\cite{Cowan2019360SI} achieved a low-cost long-wave infrared panoramic photography using nine circumferential arrays lined up with small-sized long-wave infrared cameras. This stitching method requires multiple sensors and image stitching algorithms. It requires high positioning and calibration between multiple cameras. A similar design is proposed in~\cite{Lin2019ALP}, which used four visible band cameras to synthesize a low-cost portable polycamera for stereo $360^\circ$ imaging. With the example of a two-mirror pyramid panoramic cameras, \cite{Hua2007DesignAO} illustrated optimizing the pyramid's geometry and the selection and placement of imager clusters to maximize the FoV, sensor utilization, and image uniformity.
This analysis can be extended and applied to other designs based on pyramid panoramic cameras.

\subsection{Fisheye Panoramic System}
\begin{figure*}[!t]
\centering
\subfloat{\includegraphics[width=6in]{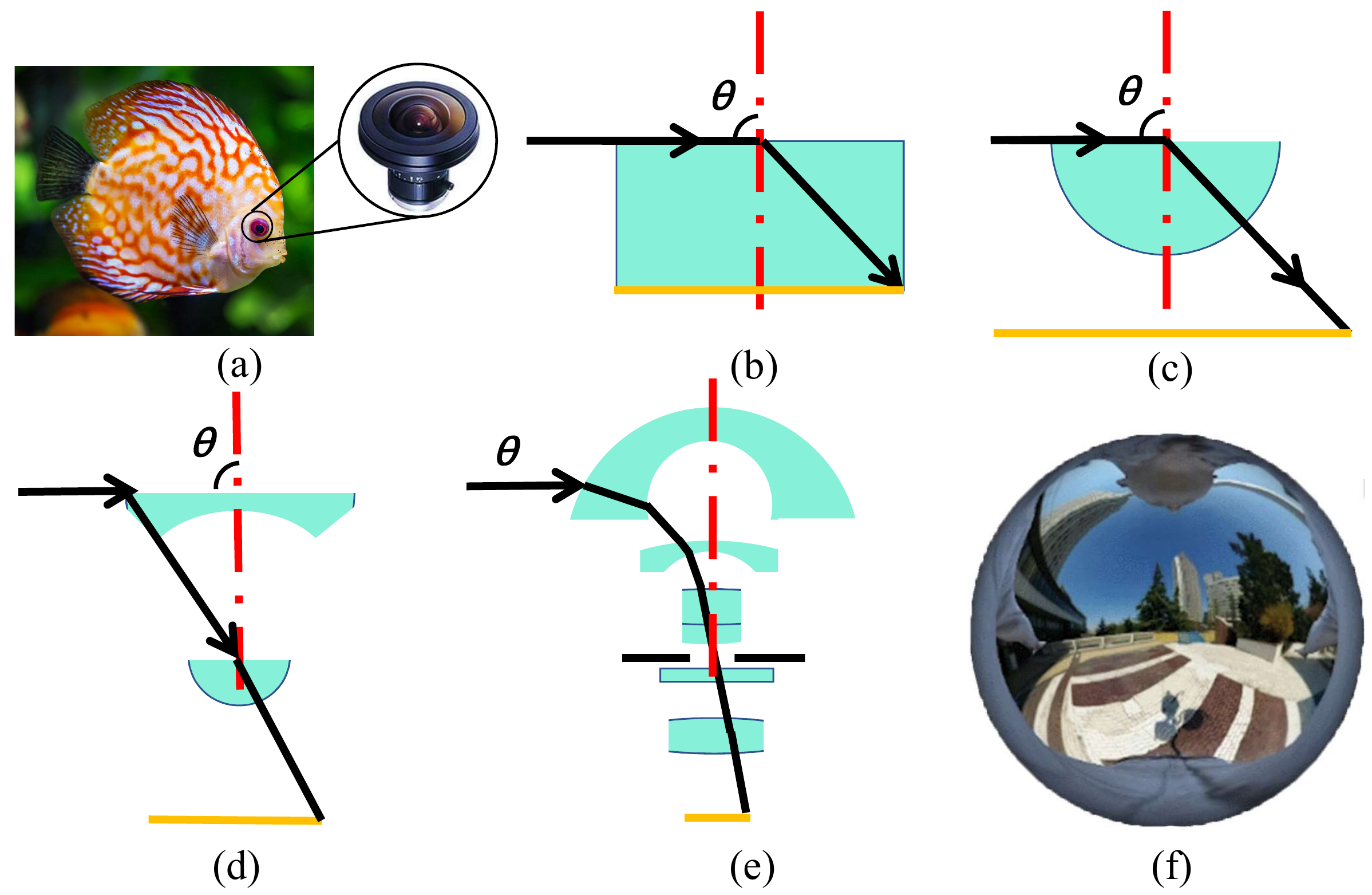}}%
\hfil
\caption{Fisheye panoramic system. (a) A fisheye camera inspired by fisheyes. Reproduced with permission~\cite{Haggui2021HumanDI}. Copyright 2021, IEEE; (b) Principle of a water-filled pinhole camera; (c) Principle of a hemispherical lens with a pupil in the center of curvature; (d) Principle of a sky lens; (e) Prototype of the modern fisheye lens. (b), (c), (d), and (e) are reproduced with permission from SPIE~\cite{Thibault2007NewGO}; (f) A panoramic image captured by a fisheye lens Entaniya M12-280. Reproduced with permission\cite{Hwang2019TowardHM}. Copyright 2019, IEEE.}
\label{fig_4}
\end{figure*}

Compared with the FoVs stitching method to obtain panoramic images, using a single camera has the advantages of simple system structure, no need for stitching algorithms, low cost, and stable installation. The most classic way is to use a fisheye optical system. It is called the fisheye optical system because its first lens is protruding and its structure is similar to a fisheye, as shown in Fig.~\ref{fig_4}(a)~\cite{Haggui2021HumanDI}.
In 1905, Wood~\cite{wood1905physical} presented a prototype of an underwater wide-angle pinhole camera in chapter four of his book \textit{Physical Optics}, as shown in Fig.~\ref{fig_4}(b). In 1922, Bond \cite{bond1922lxxxix} designed a hemispherical lens with a pupil in the center of curvature for cloud recording, as illustrated in Fig.~\ref{fig_4}(c).
In 1924, Hill~\cite{hill1924lens} proposed a hill sky lens, adding a meniscus lens before the hemispherical lens, as depicted in Fig.~\ref{fig_4}(d).
Fig.~\ref{fig_4}(e) was the first prototype of the modern fisheye lens~\cite{Thibault2007NewGO,Yan2017PhotographicZF,Thibault2008PanoramicLA,Kingslake1985DevelopmentOT,Thibault2021PanoramicLA}, patented by AEG company in 1935~\cite{AEG1935fisheye}. The FoV of the fisheye optical system usually exceeds $180^\circ$, which is an ultra-wide-angle optical system~\cite{Miyamoto:64,martin2004design}.
Due to its large FoV, it is usually used in panoramic photography and other fields. This large FoV imaging optical system generally consists of two or three negative meniscus lenses as the front group, which compresses the large FoV on the object side to the FoV required by the conventional lens, and then performs aberration correction through the relay lens group. Because the optical path of the fisheye optical system needs to be folded through multiple lenses in the front, the distortion of the optical system is large. The $f$-$\theta$ distortion is usually can be as large as $15\%{\sim}20\%$\cite{Thibault2005EnhancedOD}. The distortion control of a wide FoV lens is particularly important.
\cite{Sahin2018DistortionOF} described the design process of an imaging system for a micro wide-angle computing camera, and simulated the generation and correction method of distorted images. As shown in Fig.~\ref{fig_4}(f), a panoramic image captured by a fisheye lens Entaniya M12-280 can achieve a large-FoV imaging with no blind area of $280^{\circ}$, but there is a large barrel distortion in the edge FoV of the image~\cite{Hwang2019TowardHM}.
In addition, the first lens of fisheye optical system usually has a diameter five times greater than that of the rear correction lenses.
Relatively speaking, the compactness of the fisheye optical system caused by the large diameter difference between the front and rear lens groups is still poor. There is no blind area in the center of the fisheye systems, but the distortion of the edge FoV will compress the image. Therefore, at the beginning of the design, the compactness of the system structure, color correction, and distortion correction need to be considered~\cite{Tian2002ColourCF,Samy2015SimplifiedCF}.
For the ultra-wide-angle panoramic system, there are two kinds of aberrations of off-axis point object in each FoV: aperture-ray aberration of off-axis point object and chief ray aberration~\cite{Lu2012OptimizationMF}.
In addition to distortion correction, the correction of wave aberration of field curvature and chromatic aberration of each optical surface in fisheye optical system is also very important. The calculation method of field curvature aberration and chromatic aberration in~\cite{Fan2021CalculationOT} provided a theoretical basis for aberration correction of ultra-wide-angle systems.

In terms of image acquisition, fisheye images can be extracted by precise numerical image processing algorithms~\cite{Kweon2010ImageprocessingBP}.
Image stitching using two fisheye lenses enables a larger panoramic FoV~\cite{Ho2017DualfisheyeLS,Gao2017DualfisheyeOS,Song2020DesignOA}.
A more compact volume can be obtained by refraction of the optical path using prisms~\cite{Ye2019DesignOC,Song2018DesignAA}.
In the vacuum environment of $-40^{\circ}C{\sim}+60^{\circ}C$, the space fisheye system can still ensure stable imaging performance~\cite{Geng2017OpticalSD}. The use of capsule endoscopy to examine the pathological changes of the digestive system, especially the intestine, has recently become a major breakthrough in medical engineering.
To solve the shortcomings of the traditional endoscope capsule that the FoV is not wide enough and the imaging quality is not good enough, a novel design of micro lens with wide-angle FoV and good imaging quality is proposed in~\cite{Mang2007DesignOW}.
The system is designed with a plastic aspheric lens and a glass lens.
The length and width of the prototype are $9.8$ mm and $10.7$ mm high.
The new zoom function of fisheye optical system can be realized by using the liquid lens technology~\cite{Yen2019TheVZ}. In the field of medical and health care, the presence of blood will lead to scattering and absorption in the process of optical imaging. The fiber-optic infrared wide-angle imaging system~\cite{Peng2022DesignOA} can capture wide FoV and large depth of field infrared images in real time.

\subsection{Panomorph Imaging System}
\begin{figure*}[!t]
\centering
\subfloat{\includegraphics[width=6in]{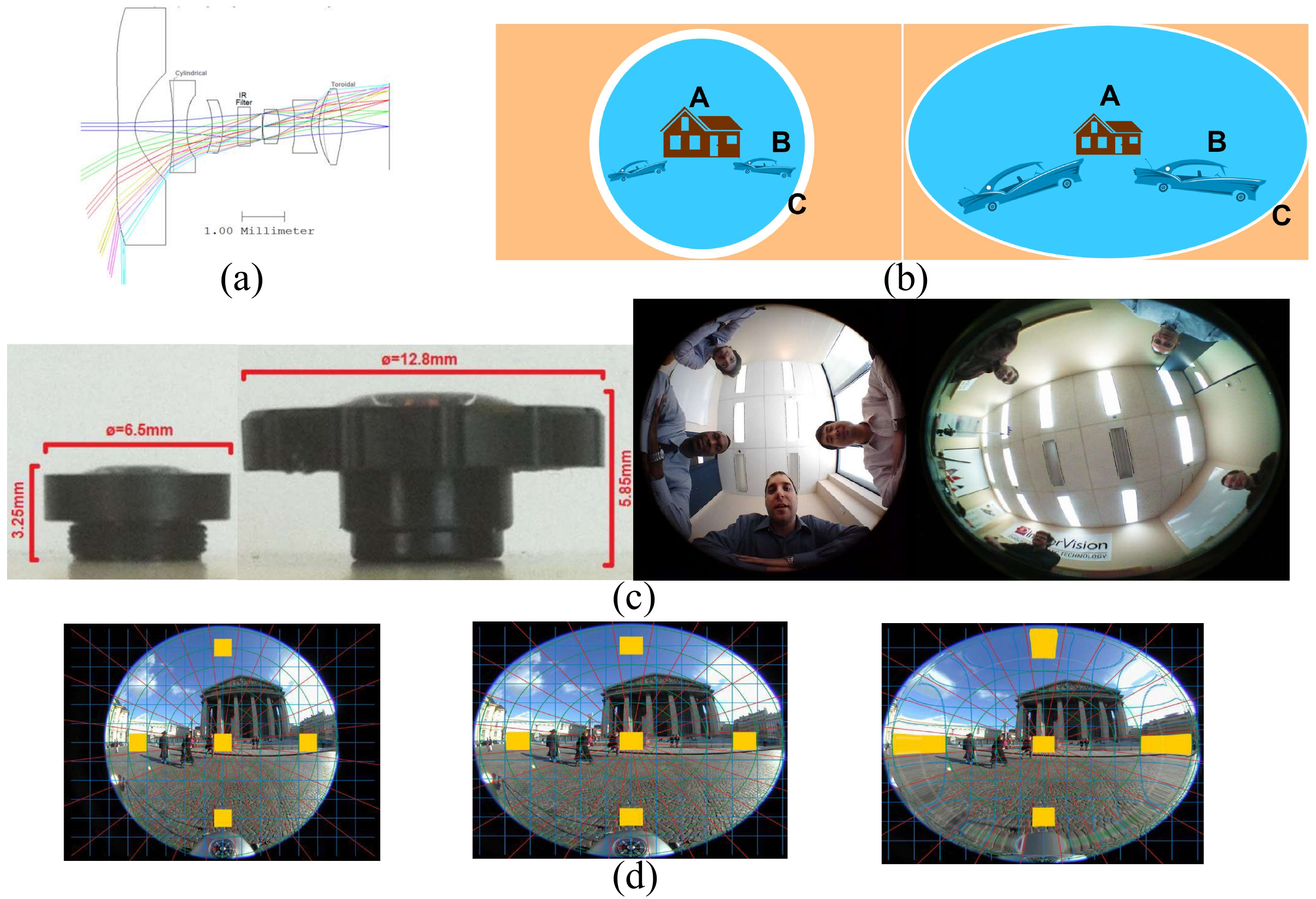}}%
\hfil
\caption{Panomorph imaging system. (a) Typical optical path of panomorph imaging system; (b) Image size distribution of fisheye lens (Left) and image size distribution of panomorph imaging system (Right). Reproduced with permission from SPIE~\cite{Thibault2007EnhancedSS}; (c) Two panomorph imaging systems (Left) and their image results (Right). (a) and (c) are reproduced with permission from SPIE~\cite{Thibault2014DesignFA}; (d) Images taken with a fisheye lens (Left) and a panomorph lens (Middle). Increased resolution with a panomorph lens (Right). Reproduced with permission from SPIE~\cite{Thibault2008NovelCP}.}
\label{fig_5}
\end{figure*}

Thibault~\cite{Thibault2006PanomorphLA} proposed a novel type of panoramic camera: Panomorph lenses in 2006, and discussed the ways to control anamorphic and distortion profile.
Such systems have higher sensor area usage and more pixels in the region of interest~\cite{Thibault2006PanomorphLC,Thibault2010PanomoprhBP}. The panomorph systems are usually equipped with cylindrical and toroidal surfaces to achieve different anamorphic ratios~\cite{Thibault2009OpticalDO,Thibault2014DesignFA,Thibault2014ConsumerEO}.
The optical path of this optical system is shown in Fig.~\ref{fig_5}(a), which can be used in a surveillance system to monitor driving conditions~\cite{Thibault2014DesignFA,Thibault2014ConsumerEO}.
As shown in Fig.~\ref{fig_5}(b), the fisheye system images the distant house A, the car B, and the edge-FoV C on the sensor. The image size distribution is shown on the left.
The image taken by the panomorph system is shown in the right picture~\cite{Thibault2007EnhancedSS}. The far house A occupies a smaller area, and the car B is larger, and the edge C has a more obvious compression effect, indicating that the system has more pixels in the area of interest and is more suitable for enhanced surveillance and driving monitoring systems.
The realized panomorph systems and imaging results are shown in Fig.~\ref{fig_5}(c)\cite{Thibault2014DesignFA}.
Fig.~\ref{fig_5}(d) shows the parking lot image, taken by a fisheye lens (Left) and a panoramic lens with equidistant projection (Middle).
The yellow area is the relative size of the objects in the image. The panomorph lens provides a resolution gain by providing an anamorphic correction. The resolution on the borders is twice as high as the resolution in the center taken by a panomorph lens, also shown in Fig.~\ref{fig_5}(d)(Right)\cite{Thibault2008NovelCP}.
Tolerancing is the basic procedure of lens design.
\cite{Parent2009TolerancingPL} studied and showed how the entrance pupil changes with the FoV of these optical systems (position and size), and come to the conclusion that the distortion comes from the front surface of the lens.
Surface irregularity error is usually used to specify the manufacturing accuracy of spherical-, aspherical-, or flat surfaces.
A spatially correlated representation of irregular slope is proposed and implemented to specify surface accuracy in~\cite{Parent2010SpatialDO} and the front surface cases of fisheye and panoramic lenses are investigated in detail.
In the design of modern panoramic optical systems, distortion control is particularly important. The design and distortion control of modern panoramic systems are analyzed and discussed in~\cite{Thibault2011DevelopmentsIM}.
The optical testing of modern panoramic lenses on the market for safety and monitoring applications is reported for the first time in~\cite{PoulinGirard2012OpticalTO}. The first test is on the measurement of image mapping, especially on the reciprocal of Instantaneous
Field of View (IFoV) expressed in pixels/degrees.
The second test is to measure the MTF of the system.
All tested lenses are coupled to the same camera to measure the MTF of the system.
Therefore, any change in the front surface will greatly affect the image footprint.
Similar image effects can be achieved by using fisheye lens image and one-directional linear interpolation algorithms~\cite{Kim2016FisheyeLC}.
To improve the relative illuminance of panomorph lenses, it is necessary to study their 3D entrance pupil mode~\cite{Zhuang2017NumericalIO}.
It is shown that with the increase of FoV and pupil size, the entrance pupil moves forward, the optical axis is shifted, tilted, and deformed~\cite{Zhuang2017AnalysisOT}.
This increased-resolution panomorph lens is particularly suitable for use in implementing vehicle positioning and tracking~\cite{Akdemir2018RealtimeVL}.
\cite{Roulet2010360EU} demonstrated that wide-angle FoV, enhanced resolution, close focusing, and distortion-free multivisualization software can improve laparoscopic and other endoscopic procedures. Furthermore, infrared panoramic lens can play a huge role in military short-distance positioning and urban security surveillance tasks~\cite{Thibault2008IRPL,Thibault2009OpticalDO}.

\subsection{Catadioptric Panoramic System}
\begin{figure*}[!t]
\centering
\subfloat{\includegraphics[width=6in]{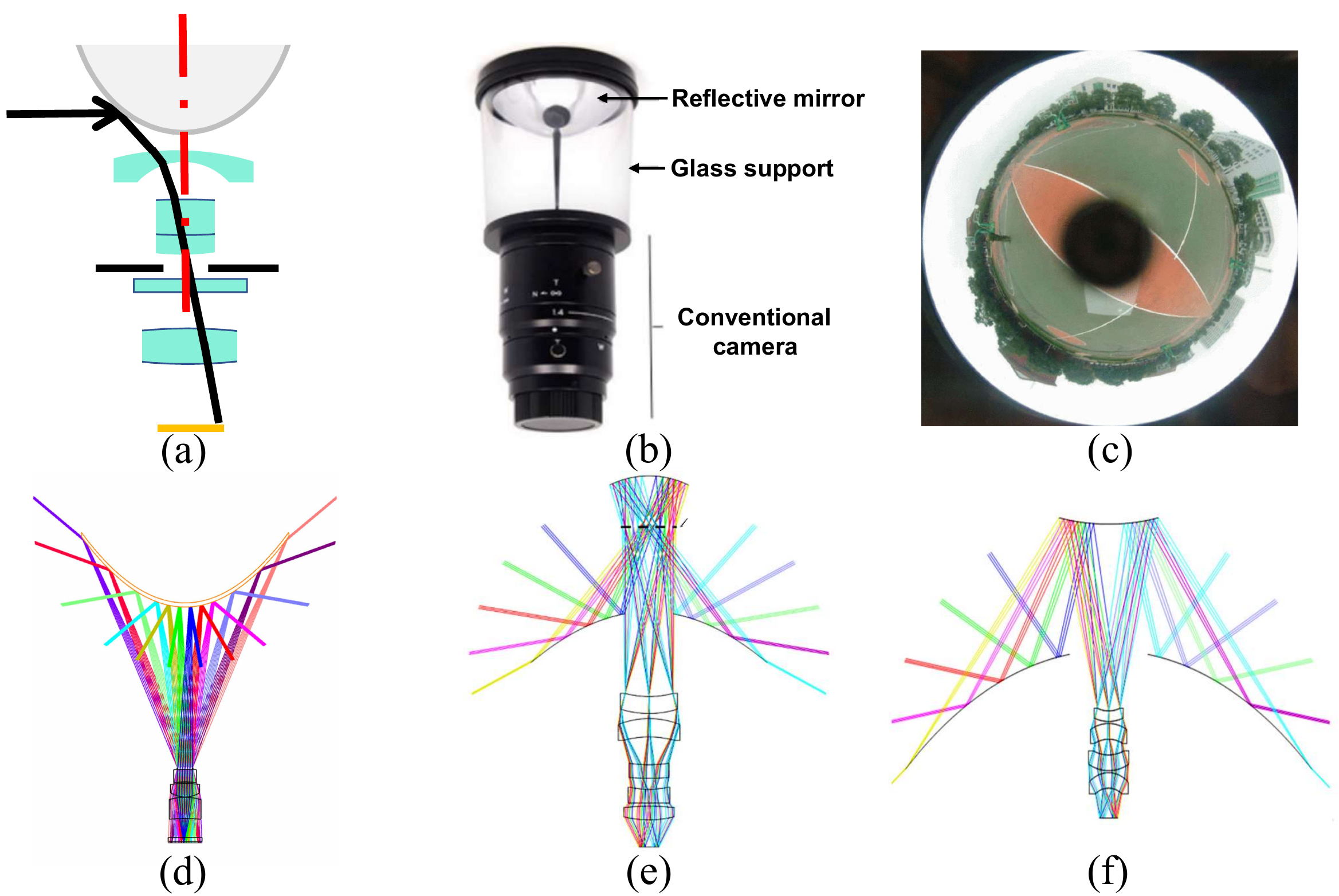}}
\hfil
\caption{Catadioptric panoramic system. (a) Imaging principle of catadioptric panoramic system; (b)  A single mirror catadioptric panoramic system. Reproduced with permission~\cite{Aziz2016ColormetricTF}. Copyright 2016, IEEE; (c) Image captured by a single mirror catadioptric panoramic system. Reproduced with permission~\cite{Wei2006ACV}. Copyright 2006, IEEE; (d) Optical path diagram of a catadioptric panoramic system with a single mirror; (e) An infrared catadioptric panoramic system with double mirrors bent in same direction; (f) An infrared catadioptric panoramic system with double mirrors bent in opposite directions. (e) and (f) are reproduced with permission from~\cite{2019Design}. Copyright 2019, INFRARED.}
\label{fig_6}
\end{figure*}

Catadioptric panoramic system is mainly composed of two parts, one part is the reflective optical element, and the other part is the refractive optical element~\cite{Chahl1997ReflectiveSF,Benosman2001PanoramicV,Kweon2005FoldedCP,Wang2022SingleviewCO}.
The reflective element is generally the mirror of the front element of the catadioptric system (Fig.~\ref{fig_6}(a)). The fisheye optical system uses multiple negative meniscus lenses for refraction, compressing the direction of the light entering relay lens group. Different from the fisheye optical system, the catadioptric panoramic optical system uses a mirror to reflect the surrounding $360^\circ$ light into the relay lens group. The refractive element of the catadioptric panoramic optical system is the relay lens group, which is used to correct aberrations for imaging.
The mirror is placed in front of the aperture diaphragm and relay lens group.
Also, the $f$-$\theta$ distortion of such panoramic optical systems is usually greater than $10\%$ due to the large FoV for imaging~\cite{Zhuang2019,2019Design}.
Compared to fisheye optics, it has fewer front lenses and focuses on surround-view imaging. Due to its characteristic as catadioptric imaging, this optical system images itself in the center of the image, which can also be considered as a blind area or area of non-interest.
Due to the inaccurate localization of feature points, the distorted chessboard image will affect the accuracy of panoramic camera calibration.
Distorted checkerboard images can affect the calibration accuracy of panoramic cameras due to the inaccurate positioning of feature points.
Compared with the conventional approach, iterative refinement method~\cite{Gong2019HighPrecisionCO} reduces the reprojection error of feature points by $39\%$.

The catadioptric panoramic system is usually equipped with a transparent glass support or center mount to connect a conventional camera and a mirror, as depicted in Fig.~\ref{fig_6}(b)~\cite{Aziz2016ColormetricTF,Aziz2020ARA}.
The center of the captured image images the sensor, which is also a kind of blind area (Fig.~\ref{fig_6}(c))~\cite{Wei2006ACV}.
A cata-fisheye camera with a $360^{\circ}{\times}55^{\circ}$ FoV can be achieved by using a fisheye camera and a mirror~\cite{Arkhipova2016CircularscanPS}.
To achieve the ideal object-image relationship, catadioptric panoramic optical systems that use a single mirror on the front element are usually designed with aspheric surfaces such as paraboloids, hyperboloids, and ellipsoids.
The corresponding area of the blind area of single mirror catadioptric panoramic system is the ground, its optical path diagram is shown in Fig.~\ref{fig_6}(d).
To achieve panoramic images in four directions with VGA video standard, a catadioptric panoramic system using an odd aspheric design is proposed in~\cite{Kweon2006DesignOA}, its image distortion is less than $0.5\%$.
Over the past few decades, the technology of image sensor provides a platform for large FoV and high-resolution 3D shooting, analysis, and static- or dynamic scenes modeling~\cite{Torii2009PanoramicA3}.
For example, vision-based driver assistance can obtain real-time 3D data from mobile platforms. The use of convex mirrors and fisheye cameras can form a panoramic stereo imaging system and can adapt to non-single-view imaging conditions for 3D reconstruction in Euclidean space~\cite{Li2011SinglecameraPS}. The concept of the catadioptric mirror can also be applied to endoscope design to simplify and reduce the volume of the system~\cite{Tseng2017PanoramicAL,Tseng2018PanoramicEB}.
The catadioptric panorama makes it possible that endoscope capsule can also use a conic mirror to observe the inner wall of the human body~\cite{OuYang2011DesignAA}.
\cite{Katkam2015CompactDE} demonstrated that a compact dual-channel endoscopic probe can simultaneously observe the needs of forward and backward FoVs in the colon.
Similar designs can also be used in endoscopic capsules~\cite{Sheu2015DualVC}.
Developing a new convex mirror and repositioning the camera viewpoint can improve the efficiency of the catadioptric camera in the RoboCup MSL robot~\cite{Lopes2011CatadioptricSO}.
This design can be used not only in the visible light band but also in a wide-spectrum infrared panoramic catadioptric system~\cite{Krishnan2008CataFisheyeCF,Spencer2006OpticalDO}.
In the military field, infrared panoramic system is also widely used~\cite{Kecskes2017LowcostPI}. The US Navy Research laboratory has developed a high-resolution medium-wave-infrared panoramic periscope sensor system~\cite{Nichols2010PerformanceCO}.
The compact system can provide $360^{\circ}$ horizontal azimuth and $-10^{\circ}{\sim}+30^{\circ}$ elevation FoV without moving components.
The design difficulty of a catadioptric imaging system mainly lies in determining the initial structural parameters of the quadric mirror of the system.
In conjunction with genetic algorithm and gradient descent, Zeng~\textit{et al.}~\cite{Zeng2020TheDO} proposed a panoramic thermal imaging system to demonstrate the effectiveness of the method.
In~\cite{Aburmad2014PanoramicTI}, the challenges involved in the three schemes to realize $360^{\circ}$ panoramic thermal imaging are discussed, and the spatial resolution, FoV, data complexity, system complexity, and cost of different solutions are analyzed in detail.
The three solutions are a $360^{\circ}$ camera design of a long-wavelength infrared XGA sensor, the splicing of three adjacent long-wave infrared sensors equipped with a low distortion $120^{\circ}$ camera, and a half-FoV $180^{\circ}$ fisheye camera with an XGA sensor.
Using a progressive design approach, Qiu~\textit{et al.}~\cite{Qiu2017OpticalDO}
designed a $14{\sim}16$ ${\upmu}$m long-wave infrared earth panorama system.
Lim~\textit{et al.}~\cite{Lim2018DesignOA} used the graphically symmetric method to correct for chromatic aberration and planar Pitzval field curvature, and designed a catadioptric system with a planar mirror. The space between the mirror surface and the relay lens group can also be changed from air to plastic~\cite{Gimkiewicz2008UltraminiatureCS,Strzl2008RuggedOM,Huang2010GeometricCO,Aikio2015OmnidirectionalLC}.
Using plastic lenses, a $360^{\circ}$ low-cost laser scanner for smart vehicle perception was designed in~\cite{Aikio2011OmnidirectionalLF}.
A front lens group can be set in front of the central FoV of the solid reflector to realize a $280^{\circ}$ catadioptric imaging system to simulate honeybee eyes~\cite{Strzl2010MimickingHE}.
The use of catadioptric optics with both forward and radial imaging channels enables minimally invasive endoscopic screening for gastrointestinal diseases, extending the detection range~\cite{Wang2011DevelopmentOA}.
To significantly improve the visual authenticity of minimally invasive surgery and diagnosis in vivo, a dual-channel panoramic endoscope is proposed in~\cite{ma2007c}, whose FoV is ${\pm}135^{\circ}$ and $360^{\circ}$ from the optical axis.
The system uses optical fiber or LED to illuminate the whole FoV.
The panoramic endoscope can prevent repeated insertion of traditional endoscopes with different perspectives and reduce the risk of misleading due to the limited FoV of traditional endoscopes. Using an annularly-stitched aspherical surface, Chen~\textit{et al.}~\cite{Cheng2016DesignOA} realized a $360^\circ{\times}270^\circ$ ultra-wide-angle three-channel catadioptric optical system. The research results show that the annularly-stitched asphere surface can significantly improve the imaging quality of different channels and realize the optimization of multi-channel and multi-FoVs imaging performance.
In~\cite{Zhou2017DoubleDC}, a double distortion correction method for panoramic images is proposed, which solves the problem of low calibration accuracy caused by the mirror distortion of the catadioptric system.
Computational ghost imaging and single-pixel imaging enable imaging and sensing under many challenging circumstances (\textit{e.g.}, scattering/turbulence, low temperature, unconventional spectroscopy). The ghost panoramic single-pixel imaging proposed in~\cite{Ye2021GhostPU} has broad application prospects in fast-moving target localization and situational awareness for autonomous driving.
 
Catadioptric panoramic imaging systems using a two-mirror design have also appeared. To reduce the blind area ratio, a catadioptric panoramic system using two mirrors with the same bending direction was proposed in~\cite{Zhang2020DesignOA}.
The system realized that the blind area is less than $3.67\%$.
Yang~\textit{et al.}~\cite{Zijian2015DesignOL} adopted a similar structure and designed a catadioptric panoramic system for imaging at $0.4{\sim}0.9$ ${\upmu}$m.
This catadioptric panoramic architecture can be used to design low F-number long-wave infrared panorama systems~\cite{Furxhi2018DesignDA}.
The design and performance comparison of the two mirrors in the same (Fig.~\ref{fig_6}(e)) and opposite (Fig.~\ref{fig_6}(f)) directions is discussed in detail in~\cite{2019Design}.
In the field of automatic driving, the use of visible and infrared dual bands can help to realize day- and night panoramic scene detection~\cite{Zhang2020ANC}.

Ultraviolet detection technology is concentrated in the $240{\sim}280$ nm ultraviolet band.
The solar radiation in this band is strongly absorbed by the ozone layer and hardly exists in the near-earth atmosphere.
It is called sun blind ultraviolet.
In the solar-blind area, the ultraviolet radiation generated by targets such as flames and high-voltage discharge coronas is easily revealed under the weak radiation background noise.
The ultraviolet detection technology realizes the monitoring of dangerous targets by detecting these spectral signals. The front element of the system can be constructed by using two mirrors and a transmissive meniscus lens to realize target detection in the ultraviolet spectrum~\cite{Liping2011DesignOC}. Panoramic detection in both visible and ultraviolet wavelengths can be achieved by using a beam splitter~\cite{Liping2010OpticalDF}. Using two Charge-Coupled Device (CCD) cameras and two parabolic convex mirrors, combined with an enhancement of existing deep feature matching methods with epipolar constraints, a practical panoramic stereo sensor can be realized~\cite{CrdovaEsparza2020ThreeDimensionalRO}.

Further, in~\cite{Ju2018OpticalDO,Ju2020OpticalDO}, the researchers designed a mirror panoramic system composed of four mirrors. It is composed of pseudo-Cassegrain collecting mirror and a reverse pseudo-Cassegrain imaging mirror, which can image visible light and long-wave infrared light simultaneously.
The FoV of the system is $360^\circ{\times}(40^\circ{\sim}110^{\circ})$ and the F-number is $1.55$.

\begin{figure*}[!t]
\centering
\subfloat{\includegraphics[width=6in]{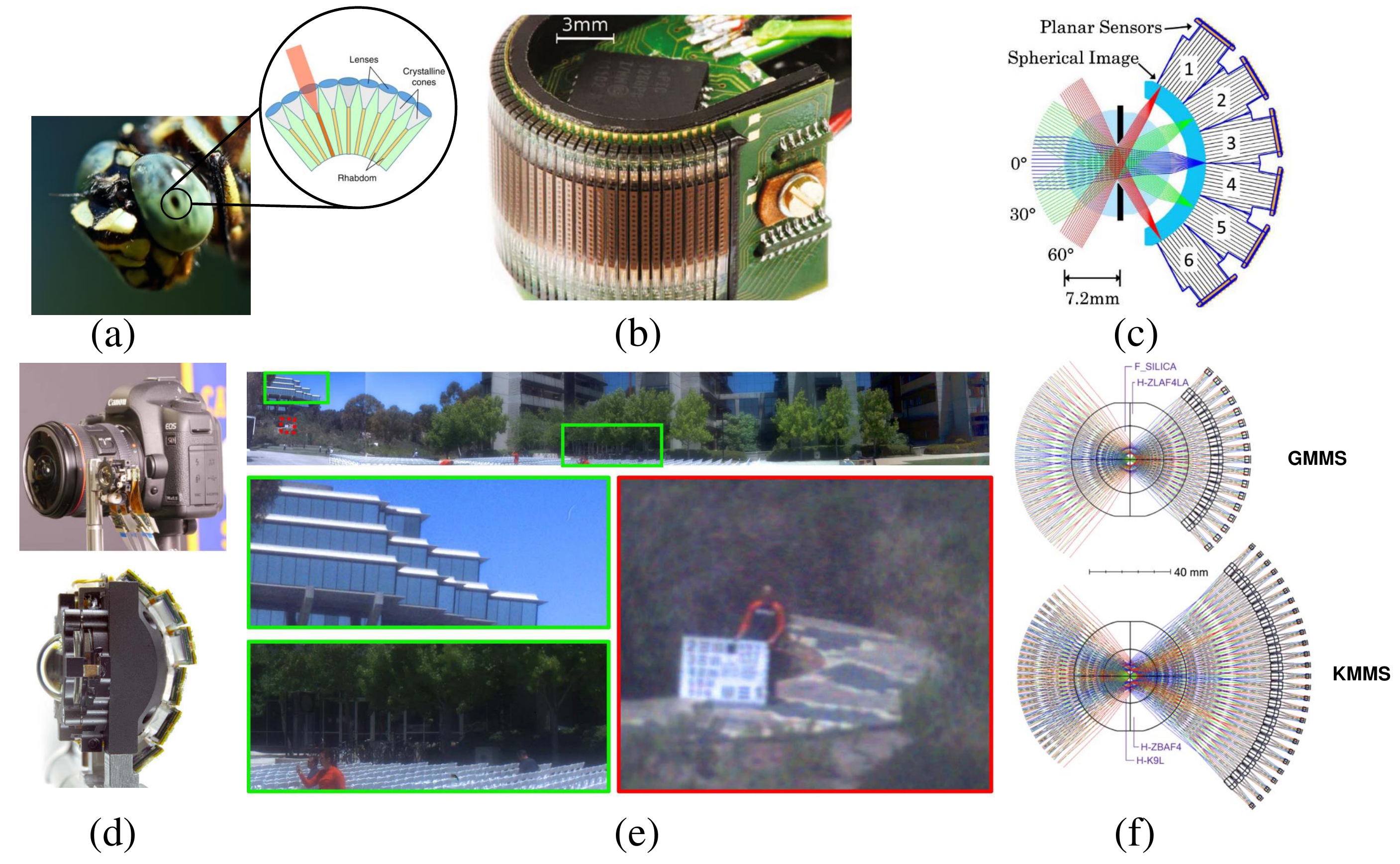}}
\hfil
\caption{Monocentric panoramic system. (a) Typical natural insect compound eye (Dragonfly). Adapted from~\cite{Kogos2020PlasmonicOF} under the Creative Commons Attribution 4.0 License; (b) CurvACE prototype. Adapted from~\cite{Viollet2014HardwareAA} under the Creative Commons Attribution 4.0 License; (c) A monocentric lens with fiber bundles to relay the image to flat sensors; (d) F/1.35, 30 megapixel, 126$^\circ$ FoV fiber-coupled monocentric lens imager prototype and similar FoV commercial F/4 camera size comparison. (e) Processed fiber relayed prototype image. (c), (d), and (e) are adapted with permission from~\cite{Olivas2015ImagePF}. Copyright 2015, Optical Society of America; (f) Optical path diagram of a GMMS (Top) system and a KMMS (Bottom) system. Adapted with permission from~\cite{Huang2020ModelingAA}. Copyright 2020, Optical Society of America.}
\label{fig_7}
\end{figure*}

\subsection{Monocentric Panoramic System}
Compound eye (Fig.~\ref{fig_7}(a))~\cite{Kogos2020PlasmonicOF,Kinoshita2005ThinCE,Ueno2013CompoundEyeCM,Wu2017ArtificialCE,Cheng2019ReviewOS,Liang2019OnelensCU,Sato2021DesignOA,zhang2022BioinspiredSC} is a typical feature of arthropod in nature.
Compared with a single aperture vision system, the compound eye possesses many excellent imaging characteristics, such as compact size, wide FoV, great perception of moving objects and sensitivity to light intensity~\cite{Phan2021ArtificialCE}.
Benefiting from the development of electronic sensors and advances in micro-nano fabrication, the realization of curved sensors makes traditional optical systems no longer limited to imaging on flat sensors.
In 2014, Viollet~\textit{et al.}~\cite{Viollet2014HardwareAA} proposed a Curved Artificial Compound Eye (CurvACE) (Fig.~\ref{fig_7}(b)) inspired by drosophila.
It is composed of a microlens array, a neuromorphic photodetector array, and a flexible Printed Circuie Board (PCB).
It consists of $630$ ommatidia and CurvACE's FoV is $180^\circ{\times}60^\circ$.
To satisfy the requirements of miniaturization, high imaging quality, large FoV of monitoring camera, and the development of curved image sensor, Wang~\textit{et al.}~\cite{wang2019design} designed a monocentric monitoring lens optical system.
The monocentric system has a FoV of $140^\circ$, $7.88$ mm focal length, $14.47$ mm total length, and F/$1.5$.
A monocentric lens with multi-aperture integration was proposed (Fig.~\ref{fig_7}(c))\cite{Olivas2015ImagePF} in 2015. This optical system has the characteristics of all optical surfaces are spherical and share a common center of curvature.
The monocentric lens with compact imager volume has no coma or astigmatic aberration.
A F/$1.35$, $30$ megapixels, $126^\circ$ fiber-coupled monocentric lens imager prototype greatly reduces the volume of the wide-FoV imaging system compared to a commercial F/$4$ camera (Fig.~\ref{fig_7}(d))~\cite{Olivas2015ImagePF}. The image processing methodology can significantly improve the quality of the fiber relayed prototype image, as illustrated in Fig.~\ref{fig_7}(e)~\cite{Olivas2015ImagePF}.
A gigapixel monocentric multiscale imager has been shown to integrate a two-dimensional mosaic of subimages~\cite{Tremblay2012DesignAS,Son2011AMW,Brady2012MultiscaleGP}. 
This wide- and uniform field can also be realized by waveguide method~\cite{Stamenov2012OptimizationOT}.
Instead of relay optics, one or more multimode fiber bundles~\cite{Ford2013SystemOO,Ford2015PanoramicIV} can be used to transfer the spherical image surface to a conventional planar image sensor.
To help determine the best specific design for a 4GA-8 lens, a wide FoV monocentric lens is studied from simple structures to moderately complex in~\cite{Stamenov2013OptimizationOH}, and a system optimization method for global optimization of dual-glass lenses are proposed for a series of large FoV imaging systems to offer practical high-performance options. The monocentric lens has recently shifted from the initial historical curiosity~\cite{kingslake1989history} to a promising approach for panoramic high-resolution imagers.
In~\cite{Stamenov2013CapabilitiesOM,stamenov2014panoramic}, the technique of monocentric lens using analytical aberration calculation and constrained numerical optimization was introduced, and its application in fully symmetric and hemispherical symmetric lenses was demonstrated.
The performance summary of the monocentric lenses in the visible and near-infrared spectral bands can be used to guide the design of various monocentric lenses.
\cite{Stamenov2014PanoramicMI} established the potential of a fiber-coupled monocentric imager in the novel  panoramic high-resolution imaging and verified the ability of the monocentric lens to focus large FoV images at a series of object depths.
By changing the fiber bundle structure can improve spatial resolution. Thanks to digital image processing, imaging quality can be further enhanced.
In~\cite{stamenov2014broad}, the visible light and visible near-infrared spectrum monocentric system which transmits the curved image to the Complementary Metal Oxide Semiconductor (CMOS) focal plane through the fiber bundle to generate the high-resolution panoramic image was described.
By eliminating the overlapping requirements of adjacent sub-images, the monocentric multi-scale system volume was reduced and the relative illumination and imaging quality were improved~\cite{Pang2017GalileanMM}.
Compared with the previous first generation of monocentric multi-scale gigabit pixel cameras with Keplerian design, the total volume of Galilean monocentric multi-scale system can be reduced by $10$ times. In~\cite{Schuster2017FoldedMI}, a novel architecture was defined for near monocentric imaging.
The folding architecture provides a more miniaturized structure, especially when the structure includes auxiliary folding mirrors.
In~\cite{Liu2019DesignOM}, researchers applied computational imaging theory to optical system design and developed a geometric aberration optimization function, which avoids using repeated iteration of optical system to define initial value of required system.
It helped to improve the time efficiency of the system design.
This design not only solved the contradiction between wide FoV and high resolution but also provided a new design idea for computational imaging.
A new analytical model of Galileo Monocentric Multi-Scale (GMMS) and Keplerian Monocentric Multi-Scale (KMMS) was proposed in~\cite{Huang2020ModelingAA}, which is more accurate than the paraxial form.
The model avoids the laborious analysis of different monocentric lens forms and maintains the key points of the Monocentric Multi-Scale (MMS) system. In addition, the correlation between GMMS and KMMS system design parameters was discussed in detail.
The results showed that the GMMS system behaves better in aberration performance.
The research provided a useful reference for the further application and development of monocentric lens system. The light field camera can still focus accurately and take clear photos under the condition of low-light and high-speed image movement.
The concept of monocentric panoramic light field is proposed by combining a monocentric lens and light field sensors in~\cite{schuster2019panoramic}. The latest research showed that the system using the combination of monocentric lens and microlens array can significantly improve the imaging quality and light field reconstruction using the calculated point spread function model~\cite{Jin2021PointSF}.

\begin{figure*}[!t]
\centering
\subfloat{\includegraphics[width=7in]{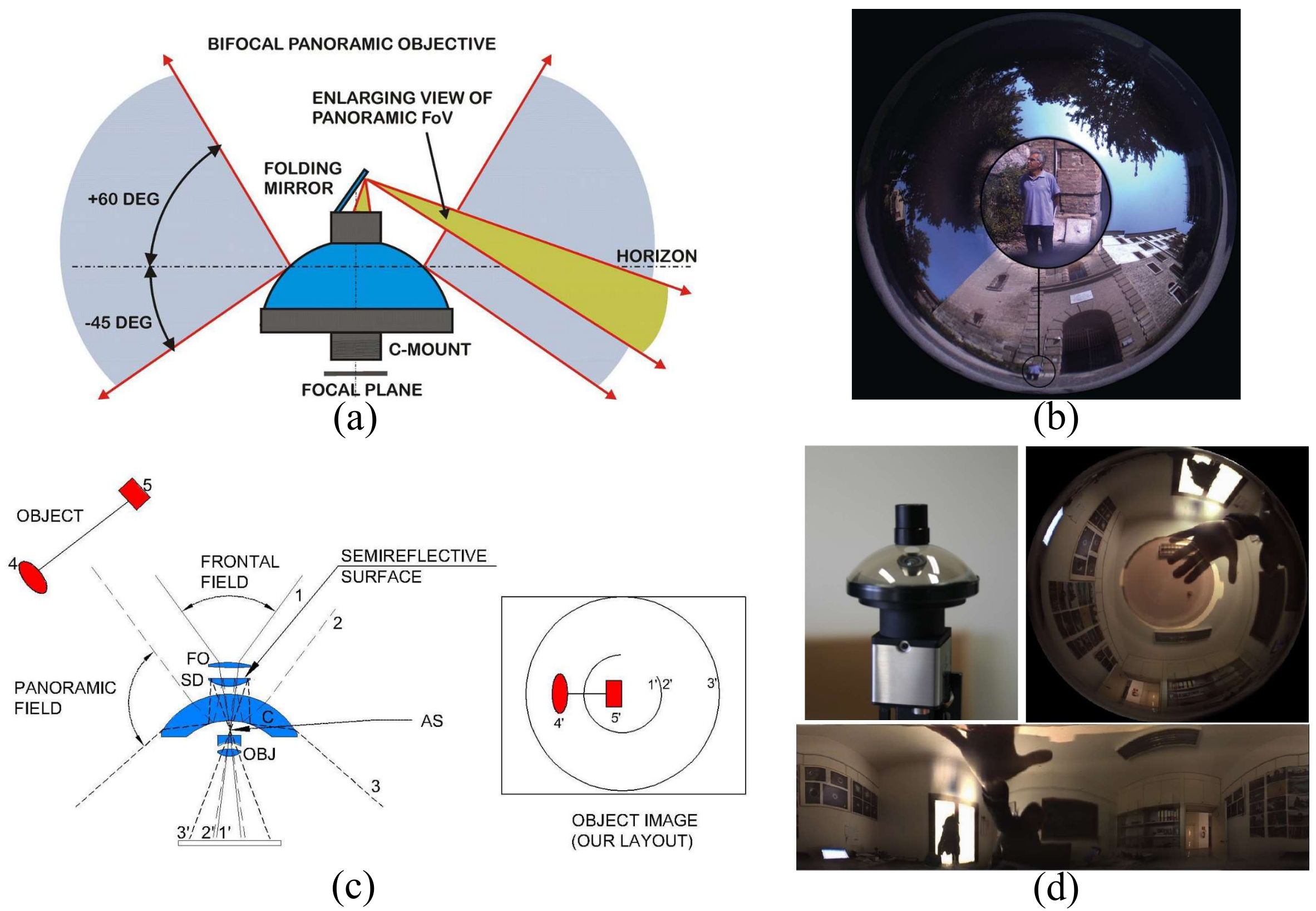}}
\hfil
\caption{Hyper-hemispheric lens. (a) A bifocal hyper-hemispheric lens; (b) An image taken with a bifocal hyper-hemispheric lens; (c) Structure of dual-channel hyper-hemispheric lens system and imaging distribution of objects in dual-channel hyper-hemispheric lens. Reproduced with permission from SPIE~\cite{Pernechele2013HyperhemisphericAB}; (d) The hyper-hemispheric lens prototype (Top Left) and the original image (Top Right) recorded by the hyper-hemispheric lens. Unwrapped image (Bottom) of hyper-hemispheric lens. (a), (b), and (d) are reproduced with permission from SPIE~\cite{pernechele2018introduction}.}
\label{fig_8}
\end{figure*}

\subsection{Hyper-hemispheric Lens}
The FoV of a common panoramic camera is greater than $200^\circ$. To realize super-large FoV imaging, a hyper-hemispheric lens~\cite{Pernechele2013HyperhemisphericAB} was proposed by Pernechele. The FoV angle of this ultra-wide-angle system is $360^\circ{\times}260^\circ$.
A kind of bifocal hyper-hemispheric lens was mentioned in~\cite{pernechele2018introduction,Simioni2020GEOMETRICALCF}. The system layout of this hyper-hemispheric lens is shown in Fig.~\ref{fig_8}(a) and its unwrapped image is shown in Fig.~\ref{fig_8}(b).
The panoramic channel uses a meniscus lens to collect and image $360^\circ$ of the surrounding light. There is a blind area in the center of the system. To avoid the waste of the imaging area of the sensor in the central blind area, the front lens group and the mirror can be used for imaging, and the imaging area can be regarded as the area of interest. The entire system has two focal lengths for the front channel and the panoramic channel.
Using an even aspheric surface, Gong~\textit{et al.}~\cite{Gong2015DesignOA} used a similar structure to design a panoramic imaging system with a full FoV of $360^\circ{\times}200^\circ$ without blind area.
Another design concept of the dual-channel panoramic system was mentioned in~\cite{Pernechele2013HyperhemisphericAB}.
By coating semi-reflective film, continuous imaging of front FoV and panoramic FoV can be realized.
As shown in Fig.~\ref{fig_8}(c), the oval and square red targets and their connecting lines can form a continuous image in the image plane of the sensor.
In~\cite{pernechele2018introduction,Pernechele2016HyperHL}, a hyper-hemispheric lens (Fig.~\ref{fig_8}(d)) is capable of imaging the FoV with an azimuth angle of $360^\circ$ and a zenithal angle of $270^\circ$ was proposed. In this study, the theoretical optical quality, projection, and angular resolution of the hyper-hemispheric lens with a paraxial focal length of $2$ mm were discussed and analyzed in detail. The example pictures of the displayed hyper-hemispheric image proved that the design can successfully obtain the panoramic image with a ultra large FoV.
The hyper-hemispheric lens can also be used as a multi-purpose panoramic camera for a star tracker to capture images.
The attitude determination algorithm of space platform was proposed in~\cite{Opromolla2017ANS}. The most advanced star tracker can accurately image as many stars as possible in a narrow- or moderate FoV, but its observation ability is limited by the FoV characteristics of the optical system.
By combining algorithmic concepts from the fields of computer vision and robotics researches, such as template matching and point cloud registration, the method was used for star recognition.
The system provides a stable and trustworthy initial attitude approach for space platforms such as satellites. It can estimate the attitude with an accuracy better than $1^\circ$ and evaluate the success rate of about $98\%$.

\begin{figure*}[!t]
\centering
\subfloat{\includegraphics[width=6.5in]{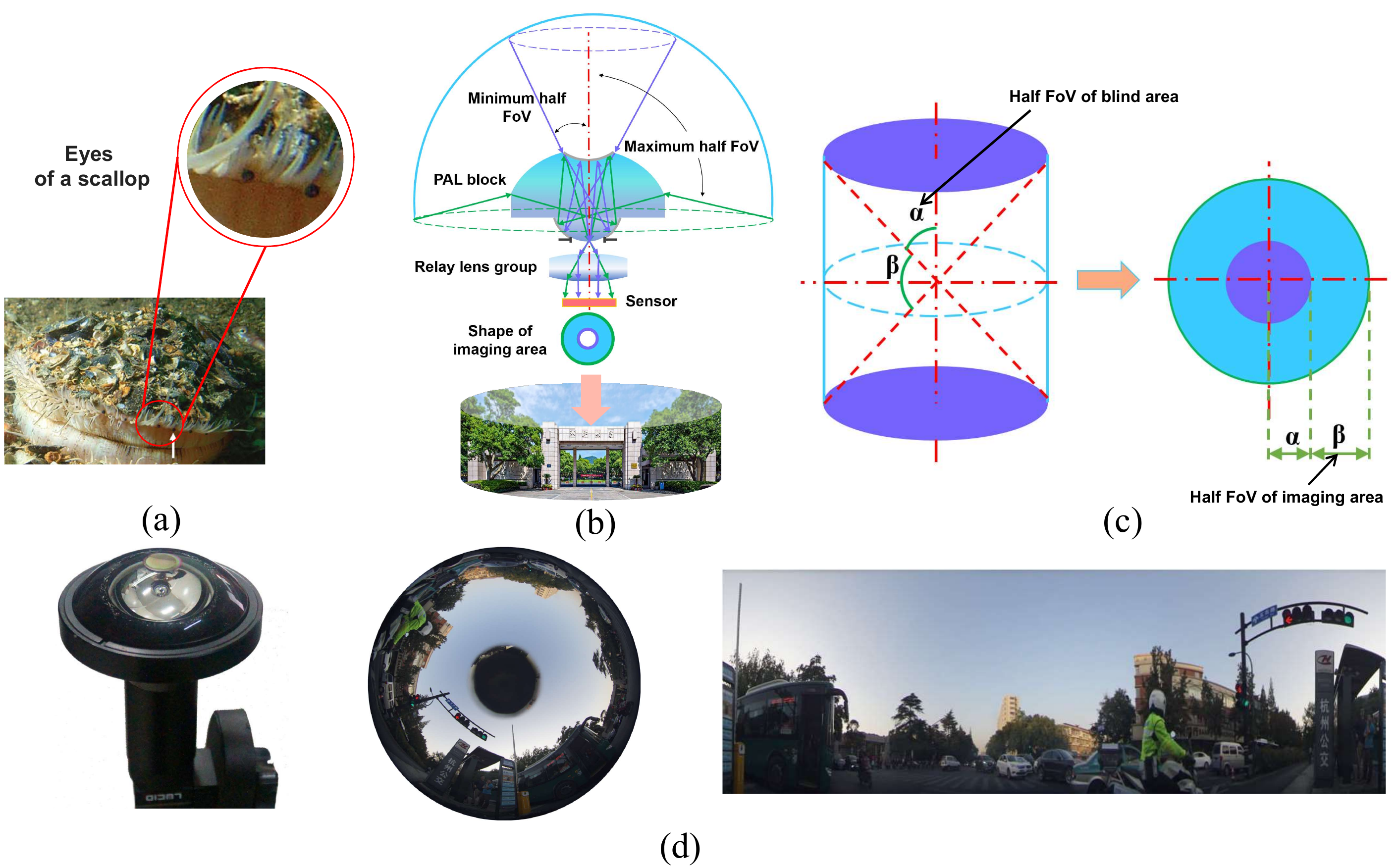}}
\hfil
\caption{Panoramic annular lens. (a) A scallop with more than 200 eyes. Reproduced with permission~\cite{Palmer2017TheIM}. Copyright 2017, The American Association for the Advancement of Science; (b) Imaging principle and composition of panoramic annular system; (c) Flat cylinder projection principle; (d) Prototype of panoramic annular imaging system (Left), raw image captured by the panoramic annular imaging system (Middle), and unwrapped image of raw image captured by the panoramic annular imaging system (Right). Reproduced with permission~\cite{yang2019pass}. Copyright 2020, IEEE.}
\label{fig_9}
\end{figure*}

\subsection{Panoramic Annular Lens}
Nature provides a steady stream of inspiration for human innovations in bionic optics.
In nature, a scallop usually possesses a visual system comprising up to 200 eyes that allow them to observe the surroundings through slits in the shell (Fig.~\ref{fig_9}(a))\cite{Palmer2017TheIM}. Each eye of the Pecten scallop consists of a highly unusual double-layered retina, a cornea, a weakly refractive lens, and a concave mirror. 
 Inspired by the distinctive imaging light path of refraction and reflection in the Pecten scallop eye\cite{greguss1991panoramic}, panoramic annular lens was first formally proposed by Greguss in 1986~\cite{greguss1986panoramic}.
The whole system is mainly composed of three parts: a panoramic annular lens block, a relay lens group and a sensor, as shown in Fig.~\ref{fig_9}(b).
Different from the traditional fisheye lens that refracts ambient light through multiple lenses to form a panoramic image (Fig.~\ref{2}), the panoramic annular lens is equipped with a catadioptric panoramic annular lens block to replace the lens group in the front of the fisheye lens. This catadioptric structure makes the panoramic annular lens more compact than the fisheye optical system. The $360^\circ$ ambient light around the panoramic annular system is converted into a small-angle beam after two refractions and two reflections in the panoramic annular lens block during imaging. The small-angle beam is imaged on the sensor after passing through the aperture diaphragm and the relay lens group to form a two-dimensional annular panoramic image.
According to the flat cylinder projection principle (Fig.~\ref{fig_9}(c)), the system can form a $360^\circ$ annular image with a half-FoV $\beta$, so it is called a panoramic annular lens. Due to the occlusion of the small mirror in the front part of the panoramic annular lens block, a blind area of the half-FoV $\alpha$ will be formed in the center of the FoV.

Afterwards, Powell studied and designed the infrared panoramic annular system~\cite{powell1994panoramic,Powell1996DesignSO}.
The prototype, raw image, and unwrapped image of the panoramic annular imaging system are shown in Fig.~\ref{fig_9}(d), which can realize high-definition panoramic imaging~\cite{yang2019pass}.
Using a multilayer diffractive optical element, Niu~\textit{et al.}~\cite{Niu2007DesignOA} realized a long-focal-length panoramic annular system with an effective focal length of $10.8$ mm and a diffraction efficiency greater than $99.3\%$.
The designed panoramic annular system can be used to monitor cavity pipes, water wells, and mines, detect and record the rapid process of the rocket launch and explosion, spacecraft orientation, navigation system, and many other spaces that need to be measured~\cite{Solomatin2007APV}.
Binocular stereo panoramic imaging can obtain and measure depth information~\cite{Huang2012DesignOP}.
Stray light will seriously affect the imaging quality in the panoramic optical system. The analysis and suppression of stray light in the optimization design can eliminate the influence of stray light and ensure that the captured image can still have good imaging ability in the face of strong light sources such as the sun~\cite{Huang2013StrayLA}.
\cite{Zhu2012CatadioptricLO} proposed new unwrapped and distortion correction methods to extract information from panoramic shots.
The concept of using a panoramic annular lens as an endoscope was first proposed in 1991~\cite{Matthys1991EndoscopicMU,Matthys1991EndoscopicIU}. It is proven that this ultra-wide-angle optical system can also be used in holographic interferometry~\cite{puliparambil1992panoramic}.
For a variety of use scenarios, \cite{Hui2012DesignsFH} described the design details of a panoramic annular monitoring system, a deformation monitoring system, a mobile camera with panoramic annular accessories, and a panoramic annular endoscope system.
\cite{Liu2016DesignOH} presented a panoramic endoscope imaging system based on the freeform surface. The front channel of the system uses a steering mirror, so that the front channel can enlarge the local details of the panoramic channel.
Compared with the traditional wide FoV endoscope, this design does not need to rotate and uses the dual-channel FoV to realize panoramic observation and local magnification, which can reduce the diagnosis time and improve the lesion detection rate.
In 2015, Wang~\textit{et al.}~\cite{Wang2015DesignOP} applied ogive surface to the design of panoramic annular lens for the first time and obtained a panoramic annular system with an effective focal length of $10.375$ mm. 

To obtain more visual field information while keeping the system compact, Huang~\textit{et al.}~\cite{Huang2017DesignOA} proposed an even ogive surface and a compact dual-channel panoramic optical system with FoVs of $360^\circ{\times}(38^\circ{\sim}80^\circ)$ and $360^\circ{\times}(102^\circ{\sim}140^\circ)$ in 2017.
In 2020, a kind of ogive panoramic annular lens was fabricated~\cite{Zhou2020DesignAI}. It realized the ultra-low distortion design with a FoV of $360^\circ{\times}(40^\circ{\sim}95^\circ)$ and $f$-$\theta$ distortion less than $0.8\%$.
This ultra-low distortion design is especially suitable for computer vision tasks such as Simultaneous Localization and Mapping (SLAM)~\cite{wang2022pal_slam}.
In terms of light field display, the pure horizontal parallax light field display of a panoramic camera was adopted to determine that it can provide smooth and continuous full parallax presentation for multiple viewers within panoramic FoV~\cite{Su2016360D}.
Q\_{bfs} aspheres can significantly reduce distortion in the design of panoramic annular lens~\cite{Zhou2014DistortionCF,zhou2016comparison_two_panoramic_front}.

Star tracker plays an essential role in satellite navigation. Star trackers on satellites in low earth orbit usually have two optical systems: one is to observe the contour of the earth and the other is to capture the position of stars. In~\cite{Liu2016DesignOH}, a star tracker optical observation system with a dual-channel panoramic annular system based on a dichroic filter was shown. It can observe the earth with a panoramic FoV range of $350{\sim}360$ nm band of $40^\circ{\sim}60^\circ$, and the front channel can capture stars far away from the system with a FoV range of $500{\sim}800$ nm band of $0^\circ{\sim}20^\circ$. In the aerospace field, the panoramic annular lens can be used to expand the FoV of astronauts in performing extravehicular activity~\cite{2018Research}. The study~\cite{Huang2015AnalysisOT} showed that the conic conformal dome of the aircraft has no obvious impact on the panoramic annular lens, and can even improve the imaging quality. It indicates that the panoramic annular lens is also suitable to be equipped in an optical window with conic conformal dome. The dual-channel panoramic annular lens structure can be used as an atmospheric detection optical system~\cite{Wang2017OpticalSD}.
Compared with the traditional optical system, the optical system can observe the sky bottom view and panoramic side view at the same time, which improves the efficiency and accuracy of observation data inversion. The imaging spectrometer of traditional atmospheric detection usually uses a prism or grating to split the beam, and can only explore the atmosphere through single-mode single detection. Using the electronic filter developed based on the crystal acoustooptic effect, a novel panoramic imaging spectrometer~\cite{Wang2019NadirAO} for atmospheric detection was designed. It can detect the hyperspectral atmospheric data of $10^\circ$ nadir FoV, $360^\circ$ limb FoV, and $0{\sim}100$ km limb height. 

The use of binary diffractive surfaces in the design process of the panoramic annular lens is beneficial for improving the imaging effect and simplifying the optical structure~\cite{2019Refractive}. To realize the design of super large FoV without a blind area, Wang~\textit{et al.}~\cite{Wang2019ExtremelyWL} proposed a dual-channel panoramic system without a blind area. The FoV of the whole system is $360^\circ{\times}230^\circ$. The FoV of the front channel is $360^\circ{\times}(0^\circ{\sim}56^\circ)$, and the FoV of the panoramic channel is $360^\circ{\times}(55^\circ{\sim}115^\circ)$. More optimization parameters can be obtained by using even aspheric surfaces, and the imaging performance of a larger FoV can be improved~\cite{Amani2020DualviewCP}.

With the rapid development of sensor technology and the emergence of polarization sensor~\cite{Gruev2011HighRC}, panoramic perception is no longer limited to visible and infrared bands. Polarization sensing can make the traditional panoramic optical system obtain higher dimensional information and realize panoramic multidimensional sensing.
An effective way to detect pavement conditions is to use polarization characteristics. The use of a panoramic annular camera and a polarization filter can capture vertical and horizontal polarization images at the same time without camera calibration~\cite{Horita2009OmnidirectionalPI}.
The low spatial resolution of elevation angle of panoramic annular image in the vertical direction will lead to quality degradation of unfolded panoramic image. Using multiple panoramic annular lens images can generate high quality panoramic images, which can effectively solve this drawback~\cite{Araki2010HighQP,Araki2012SubjectiveVO,Shibata2014HighQualityPI}. Using the particle filter can detect and track the target in panoramic image with high precision~\cite{Katayama2012DetectionAT}.
The multimodal vision sensor fusion technology~\cite{sun2019multimodal} used a polarization camera, a stereo camera, and a panoramic annular lens to realize multidimensional environment perception for autonomous driving.

With the development of information technology (such as 5G and 6G), the high-resolution panoramic annular system~\cite{Wang2019DesignOH} can record and transmit more panoramic information data in real-time. With the continuous deepening of the panoramic annular system, researchers have begun to focus on the realization of panoramic detection of different FoVs. The ultimate goal is to achieve search in a large FoV and accurate target observation and detection in a small FoV.
By optimizing the azimuth characteristics of the panoramic annular lens, a two-channel panoramic annular system with double focal lengths was proposed in~\cite{Xu2016OpticalDO}. One of the channels has a smaller focal length and a wider-FoV coarse search channel and the other longer focal length and smaller-FoV channel is called the fine observation channel.
Using XY polynomials freeform surface, Ma~\textit{et al.}~\cite{Ma2011DesignOA} proposed a new zoom mechanism to achieve a panoramic zoom that does not deviate from the center of the FoV of the observation target.
Afterwards, Wang \textit{et al.}~\cite{Wang2021DesignOA} proposed a zoom panorama optical system designed by a mechanical zoom method, which achieved a focal length change of $3.8{\sim}6$ mm and a maximum half FoV ranging from $65^\circ$ to $100^\circ$. The common zoom panoramic annular lens is based on the axial displacement between optical lens groups, which will enlarge the blind area. The zoom panoramic annular lens proposed in~\cite{Luo2014DesignOV} is based on the Alvarez freedom surface of transverse displacement, which can enlarge the imaging area and keep the blind area unchanged.
To improve design freedom degree and correct the color difference, Liu~\textit{et al.}~\cite{Liu2019DesignOC} used a three-cemented panoramic annular block to design a panoramic annular lens with FoVs of $360^\circ{\times}(66^\circ{\sim}99^\circ)$ and  $360^\circ{\times}(94^\circ{\sim}120^\circ)$, respectively. To design a new freeform surface panoramic system, using the YOZ plane-symmetric extended polynomial freeform surface, Bian~\textit{et al.}~\cite{Bian2016ADO} offered a rectangular image field panoramic system with a FoV of $140^\circ{\times}(70^\circ{\sim}110^\circ)$.
Using annularly-stitched aspheres technology, Chen~\textit{et al.}~\cite{Chen2019DesignOA,chen2020compact} demonstrated a dual-view endoscope without blind area. The FoV of the front channel is $360^\circ{\times}(0^\circ{\sim}55^\circ)$, and the FoV of the panoramic channel is $360^\circ{\times}(55^\circ{\sim}80^\circ)$.
The blind area of the panoramic annular lens reduces the sensor utilization. To solve this problem, Luo~\textit{et al.} proposed the methods of dichroic filter~\cite{Luo2016NonblindAP} and polarizing film~\cite{Luo2017CompactPD} to eliminate the center blind area. To achieve a high-resolution panoramic system while maintaining a compact structure,~\cite{gao2021design} proposed a ray-tracing theory on the influence between the entrance pupil diameter and panoramic annular block parameters and successfully presented a high-resolution panoramic system with a compact ratio of $1.5$ without blind area. There is no vignetting in the system and two channels share the relay lens group with the same parameters. To achieve a more compact and lightweight design, a panoramic annular system based on focal power distribution theory with Petzval sum correction was presented in~\cite{Gao:22}. The whole system with only three spherical lenses achieves a large FoV of $360^\circ{\times}(25^\circ{\sim}100^\circ)$, and the enhancement of the panoramic image based on PAL-Restormer enables higher performance in various computer vision tasks.

\begin{figure*}[!t]
\centering
\subfloat{\includegraphics[width=6in]{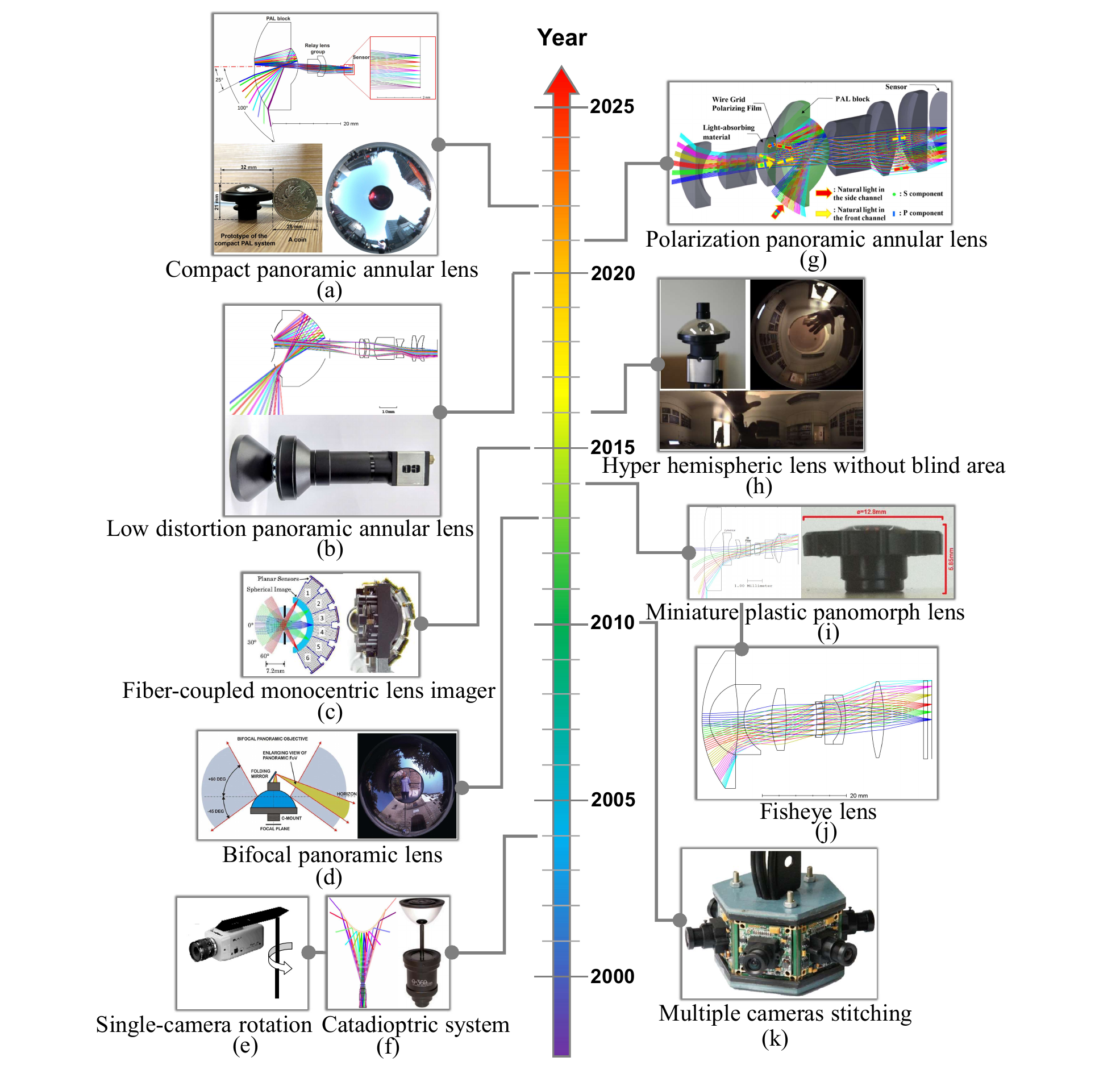}}
\hfil
\caption{Recent advances of panoramic imaging systems. (a) A compact and lightweight panoramic annular lens with a coin-size and its weight is only $20g$. Adapted with permission from~\cite{Gao:22}. Copyright 2022, Optica Publishing Group; (b) A high-performance panoramic system with $f$-$\theta$ distortion less than $0.8\%$. Adapted with permission from~\cite{Zhou2020DesignAI}. Copyright 2020, Optical Society of America; (c) A monocentric lens with fiber bundles. Adapted with permission from~\cite{Olivas2015ImagePF}. Copyright 2015, Optical Society of America; (d) A bifocal panoramic lens with enlarging view of panoramic FoV~\cite{Pernechele2013HyperhemisphericAB,pernechele2018introduction}; (e) A single camera rotation produces a panoramic image~\cite{1320265}; (f) A catadioptric panoramic system from 0-360.com~\cite{1320265}. (e) and (f) are reproduced with permission from~\cite{1320265}. Copyright 2004, IEEE; (g) A dual-channel panoramic annular lens based on polarization technology. Adapted with permission from~\cite{gao2021design}. Copyright 2021, Optical Society of America; (h) A hyper hemispheric lens with a $270^\circ$ FoV~\cite{Pernechele2016HyperHL,pernechele2018introduction}. (d) and (h) are reproduced with permission from SPIE~\cite{pernechele2018introduction}; (i) A miniature plastic panomorph lens with $180^\circ$ FoV. Reproduced with permission from SPIE~\cite{Thibault2014DesignFA}; (j) A patented $182^\circ$ FoV fisheye lens with reduced ghost reflections~\cite{ning2014wide}; (k) An optical system that uses six visible light cameras to stitch panoramic images. Reproduced with permission~\cite{Yuan2010ANM}. Copyright 2010, IEEE.}
\label{fig_Development_History}
\end{figure*}

To more clearly describe the vigorous progress of panoramic imaging systems in the last 20 years, the history of panoramic development and the comparison of key parameters of different panoramic imaging architectures are summarized in Fig.~\ref{fig_Development_History} and Table~\ref{tab:comparison_table}, respectively.

\begin{table*}[!t]
    \centering
    \caption{Comparison of key parameters among different panoramic imaging architectures.}
    \renewcommand{\arraystretch}{1.3}
\setlength\tabcolsep{0.8pt}
\begin{tabular}{c|c|c|c|c|c|c|c}
    \hline
    \multirow{2}{*}{\textbf{System category}}&\multirow{2}{*}{\textbf{Year}}&\multirow{2}{*}{\textbf{Author}}&\multicolumn{5}{c}{\textbf{Key parameters}}\\
    \cline{4-8}
    &&&\textbf{FoV}&\textbf{Wavelength}&\textbf{F-number}&\textbf{Focal length}&\textbf{Resolution/Format}\\
    \hline
    Single-camera rotation&2003&Gledhill \textit{et al.}~\cite{Gledhill2003PanoramicI}&$120^\circ$&Visible&N/A&N/A&$2048{\times}1536$\\
    \hline
    \multirow{3}{*}{Multiple cameras stitching}&2010&Yuan \textit{et al.}~\cite{Yuan2010ANM}&$360^\circ$ (6 cameras)&Visible&N/A&N/A&N/A\\
    \cline{2-8}
    &2014&Huang \textit{et al.}~\cite{Huang2014A3P}&$360^\circ$ (4 cameras)&Visible&N/A&N/A&N/A\\
    \cline{2-8}
    &2019&Cowan \textit{et al.}~\cite{Cowan2019360SI}&$360^\circ$ (9 cameras)&IR&1.1&1.4 mm&$576{\times}60$\\
    \hline
    \multirow{6}{*}{Fisheye panoramic system}&2015&Samy \textit{et al.}~\cite{Samy2015SimplifiedCF}&$160^\circ$&400$\sim$700 nm&2.8&1 mm&1/3.2 inch\\
    \cline{2-8}
    &2017&Geng \textit{et al.}~\cite{Geng2017OpticalSD}&$180^\circ$&480$\sim$650 nm&5&5 mm&$1920{\times}1080$\\
    \cline{2-8}
    &2018&Song \textit{et al.}~\cite{Song2018DesignAA}&$190^\circ$${\times}$2&423$\sim$660 nm&2&1.43 mm&$4768{\times}3516$\\
    \cline{2-8}
    &2019&Hwang \textit{et al.}~\cite{Hwang2019TowardHM}&$280^\circ$&Visible&N/A&N/A&N/A\\
    \cline{2-8}
    &2020&Song \textit{et al.}~\cite{Song2020DesignOA}&$210^\circ$${\times}$2&423$\sim$660 nm&1.8&Afocal lens&\textgreater$2000{\times}1000$\\
    \cline{2-8}
    &2021&Pernechele \textit{et al.}~\cite{Pernechele2021TelecentricFF}&$180^\circ$&500$\sim$770 nm&3&3.3 mm&$2000{\times}2000$\\

    \hline
    \multirow{3}{*}{Panomorph imaging system}&2006&Thibault~\cite{Thibault2006PanomorphLA}&$182^\circ$&400$\sim$700 nm&1.9&1.15/0.9 mm&$640{\times}480$\\
     \cline{2-8}
    &2014&Thibault \textit{et al.}~\cite{Thibault2014DesignFA}&$180^\circ$&400$\sim$700 nm&2.8&N/A&$1600{\times}1200$\\
     \cline{2-8}
    &2014&Thibault \textit{et al.}~\cite{Thibault2014DesignFA}&$180^\circ$&400$\sim$700 nm&2.8&N/A&$2592{\times}1944$\\
    
    \hline
    \multirow{8}{*}{Catadioptric panoramic system}&2006&Kweon \textit{et al.}~\cite{Kweon2006DesignOA}&$360^\circ{\times}(50^\circ{\sim}110^\circ)$&Visible&5.53&24.8  mm&$3000{\times}2200$\\
    \cline{2-8}
    &2015&Sheu \textit{et al.}~\cite{Sheu2015DualVC}&$360^\circ{\times}(0^\circ{\sim}45^\circ,130^\circ{\sim}145^\circ)$&Visible&3.5&0.76,1.23mm&$320{\times}240$\\
   \cline{2-8}
    &2016&Cheng \textit{et al.}~\cite{Cheng2016DesignOA}&$360^\circ{\times}(0^\circ{\sim}50^\circ,50^\circ{\sim}135^\circ)$&486$\sim$656 nm&2.5&1 mm&1.3 Mega,1/2.56 inch\\
    
    \cline{2-8}
    &2019&Zhuang \textit{et al.}~\cite{Zhuang2019}&$360^\circ{\times}(45^\circ{\sim}105^\circ)$&480$\sim$650 nm&2.3&0.5 mm&$2592{\times}1944$\\
    \cline{2-8}
    &2019&Wu \textit{et al.}~\cite{2019Design}&$360^\circ{\times}(40^\circ{\sim}130^\circ)$&8$\sim$12 $\upmu$m&1.2&N/A&$384{\times}288$\\
    \cline{2-8}
    &2019&Wu \textit{et al.}~\cite{2019Design}&$360^\circ{\times}(40^\circ{\sim}120^\circ)$&8$\sim$12 $\upmu$m&1.2&N/A&$384{\times}288$\\
    \cline{2-8}
    &2020&Zhang \textit{et al.}~\cite{Zhang2020DesignOA}&$360^\circ{\times}(22^\circ{\sim}120^\circ)$&486$\sim$656 nm&4&2 mm&N/A\\

    \hline
    \multirow{7}{*}{Monocentric panoramic system}&2012&Stamenov \textit{et al.}~\cite{Stamenov2012OptimizationOT}&$120^\circ$&Visible&1.7&12 mm&N/A\\
    \cline{2-8}
    &2013&Stamenov \textit{et al.}~\cite{Stamenov2013OptimizationOH}&$120^\circ$&380$\sim$550 nm&1.79&12 mm&N/A\\
    \cline{2-8}
    &2013&Stamenov \textit{et al.}~\cite{Stamenov2013OptimizationOH}&$120^\circ$&500$\sim$900 nm&1.2&16 mm&N/A\\
    \cline{2-8}
    &2013&Stamenov \textit{et al.}~\cite{Stamenov2013OptimizationOH}&$120^\circ$&900$\sim$1500 nm&1.19&12 mm&$1280{\times}1024$\\
    \cline{2-8}
    &2013&Stamenov \textit{et al.}~\cite{Stamenov2013OptimizationOH}&$120^\circ$&486$\sim$656 nm&2.33&112 mm&N/A\\
    \cline{2-8}
    &2013&Stamenov \textit{et al.}~\cite{Stamenov2013OptimizationOH}&$120^\circ$&450$\sim$700 nm&2.8&280 mm&N/A\\
    \cline{2-8}
    &2019&Wang \textit{et al.}~\cite{wang2019design}&$140^\circ$&486$\sim$656 nm&1.5&7.88 mm&11 Mega\\

    \hline
    \multirow{2}{*}{Hyper-hemispheric lens}&2013&Pernechele~\cite{Pernechele2013HyperhemisphericAB,pernechele2018introduction}&$360^\circ{\times}(0^\circ{\sim}10^\circ,30^\circ{\sim}135^\circ)$&500$\sim$650 nm&3.5&6,2mm&2/3 inch\\
    \cline{2-8}
    &2016&Pernechele~\cite{Pernechele2016HyperHL,pernechele2018introduction}&$360^\circ{\times}(0^\circ{\sim}30^\circ,30^\circ{\sim}135^\circ)$&500$\sim$650 nm&3.5&2 mm&2/3 inch\\

    \hline
    \multirow{25}{*}{Panoramic annular lens}&1994&Powell~\cite{powell1994panoramic}&$360^\circ{\times}(70^\circ{\sim}110^\circ)$&3$\sim$5 $\upmu$m&1.5&2.65 mm&$512{\times}512$\\
    \cline{2-8}
    &2007&Niu \textit{et al.}~\cite{Niu2007DesignOA}&$360^\circ{\times}(50^\circ{\sim}100^\circ)$&486$\sim$656 nm&3.7&10.8 mm&$5120{\times}4096$\\
    \cline{2-8}
    &2011&Wang \textit{et al.}~\cite{Liping2011DesignOC}&$360^\circ{\times}(45^\circ{\sim}90^\circ)$&245$\sim$285 nm&2&N/A&N/A\\
    \cline{2-8}
    &2012&Hui \textit{et al.}~\cite{Hui2012DesignsFH}&$360^\circ{\times}(52^\circ{\sim}108^\circ)$&486$\sim$656 nm&2&0.96 mm&1/3 inch\\
    \cline{2-8}
    &2012&Hui \textit{et al.}~\cite{Hui2012DesignsFH}&$360^\circ{\times}(57^\circ{\sim}100^\circ)$&Visible&2.8&0.82 mm&5 Mega\\
    \cline{2-8}
    &2012&Hui \textit{et al.}~\cite{Hui2012DesignsFH}&$360^\circ{\times}(30^\circ{\sim}90^\circ)$&Visible&2&0.27 mm&N/A\\
    \cline{2-8}
    &2012&Huang \textit{et al.}~\cite{Huang2012DesignOP}&$360^\circ{\times}(60^\circ{\sim}105^\circ)$&486$\sim$656 nm&3.18,3.22&3.19 mm&$4872{\times}3248$\\
    \cline{2-8}
    &2014&Xue \textit{et al.}~\cite{qingsheng2014optical}&$360^\circ{\times}(70.9^\circ{\sim}73.3^\circ)$&355$\sim$365 nm&3.3&5 mm&$1024{\times}1024$\\
    \cline{2-8}
    &2015&Wang \textit{et al.}~\cite{Wang2015DesignOP}&$360^\circ{\times}(45^\circ{\sim}85^\circ)$&486$\sim$656 nm&N/A&10.375 mm&N/A\\
    \cline{2-8}
    &2016&Liu \textit{et al.}~\cite{Liu2016DesignOH}&$360^\circ{\times}(0^\circ{\sim}10.2^\circ,60^\circ{\sim}97.5^\circ)$&486$\sim$656 nm&2.65,2.8&2.12,0.82mm&$648{\times}488$\\
    \cline{2-8}
    &2016&Yao \textit{et al.}~\cite{Yao2016DesignOA}&$360^\circ{\times}(30^\circ{\sim}100^\circ)$&8$\sim$12 $\upmu$m&1.15&3.4 mm&$640{\times}480$\\
 
   \cline{2-8}
    &2016&Zhou \textit{et al.}~\cite{zhou2016comparison_two_panoramic_front}&$360^\circ{\times}(30^\circ{\sim}120^\circ)$&486$\sim$656 nm&3.8&0.86 mm&1/3 inch\\
    
    \cline{2-8}
    &2017&Huang \textit{et al.}~\cite{Huang2017DesignOA}&$360^\circ{\times}(38^\circ{\sim}80^\circ,102^\circ{\sim}140^\circ)$&486$\sim$656 nm&5&2.75,3.12 mm&N/A\\
    
    \cline{2-8}
    &2017&Luo \textit{et al.}~\cite{Luo2017CompactPD}&$360^\circ{\times}(0^\circ{\sim}50^\circ,50^\circ{\sim}105^\circ)$&486$\sim$656 nm&3.3,3.0&0.833,0.895mm&$2592{\times}1944$\\
    
    \cline{2-8}
    &2019&Wang \textit{et al.}~\cite{Wang2019NadirAO}&$360^\circ{\times}(0^\circ{\sim}5^\circ,53.1^\circ{\sim}56^\circ)$&450$\sim$800 nm&6.5&7.5,4mm&N/A\\
    
    \cline{2-8}
    &2019&Wang \textit{et al.}~\cite{Wang2019ExtremelyWL}&$360^\circ{\times}(0^\circ{\sim}56^\circ,55^\circ{\sim}115^\circ)$&486$\sim$656 nm&5&3,5mm&$2048{\times}2048$\\
    
    \cline{2-8}
    &2019&Chen \textit{et al.}~\cite{Chen2019DesignOA}&$360^\circ{\times}(0^\circ{\sim}55^\circ,55^\circ{\sim}80^\circ)$&486$\sim$656 nm&3.4&0.83,1.17mm&$1280{\times}800$\\
    
    \cline{2-8}
    &2019&Wang \textit{et al.}~\cite{Wang2019DesignOH}&$360^\circ{\times}(30^\circ{\sim}100^\circ)$&486$\sim$656 nm&3.98&4.47 mm&$6000{\times}4000$\\
    
    \cline{2-8}
    &2020&Amani \textit{et al.}~\cite{Amani2020DualviewCP}&$360^\circ{\times}(0^\circ{\sim}40^\circ,40^\circ{\sim}139^\circ)$&486$\sim$656 nm&5&0.417,0.433mm&$2592{\times}1944$\\
    
    \cline{2-8}
    &2020&Zhou \textit{et al.}~\cite{Zhou2020DesignAI}&$360^\circ{\times}(40^\circ{\sim}95^\circ)$&486$\sim$656 nm&4.5&4.34 mm&$2048{\times}2048$\\
    
     \cline{2-8}
    &2021&Gao \textit{et al.}~\cite{gao2021design}&$360^\circ{\times}(0^\circ{\sim}45^\circ,45^\circ{\sim}100^\circ)$&486$\sim$656 nm&2.5&0.8,1.55mm&$4032{\times}3024$\\
    
    \cline{2-8}
    &2021&Wang \textit{et al.}~\cite{Wang2021DesignOA}&$360^\circ{\times}(25^\circ{\sim}100^\circ/25^\circ{\sim}65^\circ)$&486$\sim$656 nm&4.22${\sim}$6.51&3.8${\sim}$6mm&N/A\\
    
    \cline{2-8}
    &2021&Wang \textit{et al.}~\cite{wang2022high_performance}&$360^\circ{\times}(30^\circ{\sim}100^\circ)$&486$\sim$656 nm&4.7&3.86 mm&$5280{\times}3956$\\
    
    \cline{2-8}
    &2022&Wang \textit{et al.}~\cite{wang2022design}&$360^\circ{\times}(30^\circ{\sim}150^\circ)$&486$\sim$656 nm&3.97,2.98&1.27,4.48mm&$6252{\times}4176$\\
    
     \cline{2-8}
    &2022&Gao \textit{et al.}~\cite{Gao:22}&$360^\circ{\times}(25^\circ{\sim}100^\circ)$&486$\sim$656 nm&5.5&1.17 mm&$1280{\times}1024$\\
    
    \hline
    \end{tabular}
    \label{tab:comparison_table}
\end{table*}

\section{New Engines of Panoramic Optical Systems}
\subsection{Freeform Surface}
\begin{figure*}[!t]
\centering
\subfloat{\includegraphics[width=6.5in]{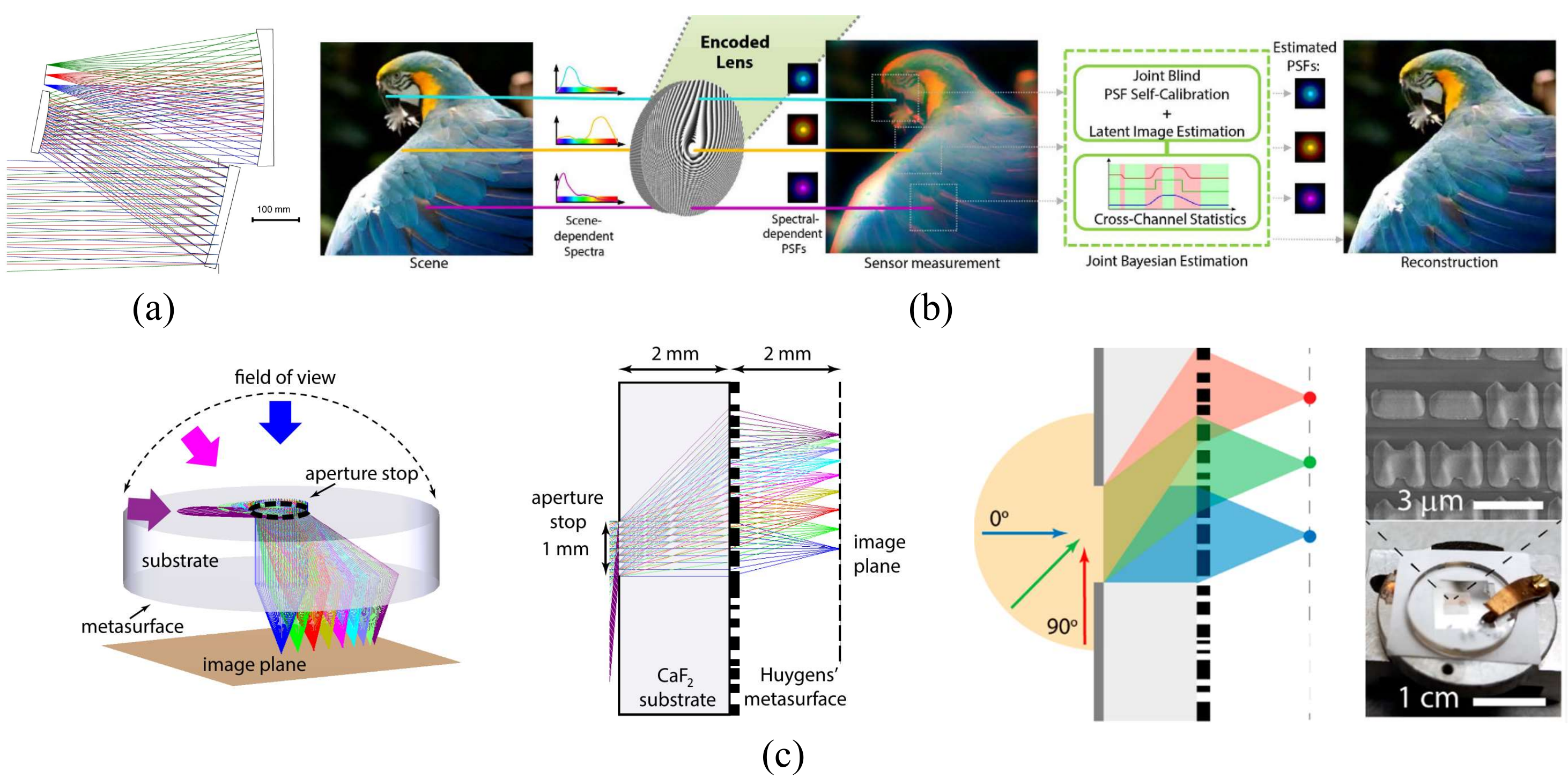}}
\hfil
\caption{New engines of panoramic optical systems. (a) A compact and high performance freeform optical system. Adapted from~\cite{bauer2018starting} under the Creative Commons Attribution 4.0 License; (b) Diffraction-encoded lenses and computational imaging techniques enable image restoration. Adapted from~\cite{heide2016encoded} under the Creative Commons
Attribution 4.0 License; (c) A single layer plane metalens with ultra-wide FoV (Left), side view of an exemplary design based on mid infrared Huygens metasurface (Middle), diffraction-limited focusing and imaging principles performed on a 180$^\circ$ FoV, metalens sample and scanning electron microscopy images of the meta-atoms (Right). Reprinted with permission from~\cite{Shalaginov2020SingleElementDF}. Copyright 2020, American Chemical Society.}
\label{fig_10}
\end{figure*}

With the continuous development of advanced manufacturing technology and testing technology, freeform optical elements make it possible for compact and high-image-quality imaging optical systems~\cite{Li2012DesignAF,Ye2017ReviewOO,YangTong2021FreeformIO,Duerr2021FreeformIS}. Optical freeform surfaces in imaging optical design generally refer to optical surfaces that do not have axis-rotational symmetry or translation-rotational symmetry~\cite{Thompson2012FreeformOS,Wills2017FreeformON,Reimers2017FreeformSE,YangTong2021FreeformIO}. Compared with the traditional spherical surface, the freeform surface can provide more degrees of freedom, significantly improve the optical performance of the imaging system, and reduce the volume of the system~\cite{bauer2018starting} (Fig.~\ref{fig_10}(a)).
This technology has significantly promoted applications from science to a wide range of fields, such as extreme ultraviolet lithography and space optics~\cite{Rolland2021FreeformOF}. Freeform optics is a concurrent engineering process that combines design, manufacturing, testing, and assembly. The continuous development of freeform optics has important significance and value for the whole human society to perceive the panoramic environment.

The traditional panoramic lens design uses spherical lens.
To improve the freedom of optical designers in designing the panoramic system, the use of rotationally symmetric aspheric surfaces can give panoramic imaging systems a new opportunity. Thanks to the multi-degree of freedom design parameters of the aspheric surface, the panoramic imaging system can realize the system parameters, structure, and functions that are difficult to be realized in traditional panoramic systems.
The proposed odd aspheric surface~\cite{Kweon2006DesignOA}, even aspheric surface~\cite{Amani2020DualviewCP}, Q-type aspheric surface~\cite{Forbes2007ShapeSF,Forbes2010RobustAF,Forbes2010RobustEC}, extended polynomial aspheric surface~\cite{Bian2016ADO}, annularly-stitched aspheric surface~\cite{Cheng2015OpticalDA,Cheng2016DesignOA}, ogive aspheric surface~\cite{Wang2015DesignOP,Zhou2020DesignAI}, and even ogive aspheric surface~\cite{Huang2017DesignOA} have been proved to be successfully applied to the design of panoramic systems.
The asymmetric freeform surface also makes the panoramic optical system have a new zoom function, including Alvarez surface freeform surface~\cite{Luo2014DesignOV} and XY polynomial aspheric surface~\cite{Ma2011DesignOA}. In the future, we believe that the new freeform surface technology can make panoramic imaging optical systems obtain more exciting new functions, high imaging quality, compact volume, and so on.

Although freeform surfaces have revolutionized the field of optical design in the last decade, there are still some challenges and some directions to be explored in this technology~\cite{YangTong2021FreeformIO}. In the design area, more powerful global optimization algorithms need to be explored to achieve fast, accurate, and generalized direct point-by-point algorithms. In terms of applications, there is an urgent need to further improve the accuracy and efficiency of freeform surfaces for fabrication, testing, and assembly. In conclusion, the freeform technology has much potential for exploration and research, and is a powerful engine for the development of panoramic imaging technology.

\subsection{Thin-plate Optics}
Thin-plate optics is a new imaging approach that uses Fresnel optics or diffractive optics or other optical elements to design thin-plate lenses that are combined with computational imaging techniques to recover or enhance images~\cite{Peng2019LearnedLF}. This technique allows the use of lightweight optical elements to build a compact optical system with the same or better image quality than conventional bulky imaging system. The thin-plate optical technology will facilitate the lightweight and miniaturization of the panoramic imaging system.

Different from traditional optical imaging, computational imaging technology~\cite{shao2020latest,Peng2015ComputationalIU,Jiang2022AnnularCI} can encode and decode optical information captured by optical instruments in all directions from information acquisition, information transmission, and information conversion in terms of imaging principles. Computational imaging can acquire and analyze multidimensional information of the light field through scattering, polarization, and bionic technologies, and has many advantages in achieving large detection distance, high resolution~\cite{Sitzmann2018EndtoendOO,Sun2020EndtoendLO}, high signal-to-noise ratio, multidimensional information~\cite{Peng2016TheDA}, light weight\cite{Qi2022AlldayTC}, simplicity, and cheapness.
For decades, optics researchers have been working to design compact optical systems with large FoV and light weight~\cite{Zhang2012RealSC,Sun2021EndtoendCL}.
A Fresnel lens~\cite{Xie2011ConcentratedSE} is an optical element that reduces the thickness of the lens while maintaining the shape of the lens curvature and can be used as a lightweight alternative to traditional continuous surface lenses. Due to its ability to reduce the thickness of the optical system, it is widely used in non-imaging fields such as illumination, solar concentrators, and collimators. When the processing of the optical lens surface is close to the light wavelength of the imaging band, the transmission of light will no longer conform to the three transmission laws of geometric optics (straight propagation, refraction, and reflection), and a diffraction effect will occur. At this time, the optical element is called diffractive optical elements. An optical system designed with diffractive elements needs to ensure that the entire optical system has a high diffraction efficiency.
Combining computational imaging technology and Fresnel/diffractive optical elements, thin-plate optics technology came into being~\cite{Peng2019LearnedLF}.
Using diffraction-coded lens to form significantly different point spread functions for different spectral distributions, joint computational imaging technology can achieve effective phase difference recovery and image reconstruction~\cite{heide2016encoded} (Fig.~\ref{fig_10}(b)).
With computational imaging technology, lightweight optical systems with large FoVs using Fresnel lenses or diffractive optical elements can achieve imaging quality close to that of traditional complex optical systems, enabling simple systems perception of large FoVs~\cite{Dun2020LearnedRS}. This next-generation imaging optical system using Fresnel or Refractive-Diffractive hybrid and computational imaging can be used to build the future of computational imaging for thin- and lightweight panoramic imaging.

With the continuous development of optical micro-nano processing technology, the processing of thin-plate optical technology has gradually matured, but there are still some challenges to be solved. In terms of principle, diffractive optical elements need to be combined with new design algorithms to improve the numerical stability and convergence speed of the design results~\cite{Huo2022History}. In terms of performance, the diffraction efficiency needs to be improved and the wavelength range needs to be further expanded. In terms of applications, the exploration of thin-plate optical technology in the field of computational imaging will be more far-reaching, expanding the direction of new panoramic intelligent perception.

\subsection{Wide-FoV Metalens}
At present, with the rapid development of optical technology and manufacturing technology, the miniaturization of optical systems has become a focus of researches.
For scene perception, wearable devices, medical devices, aerial photography, and other fields, miniaturized optical systems are favored~\cite{Li2019MetalensBasedMO,Chen2020BroadbandBL,Phan2021ArtificialCE,Chen2021PolarizationinsensitiveGM,Solntsev2021MetasurfacesFQ,Ali2021AHB}.
The miniaturization design of traditional optical system checks and balances with high resolution, high imaging quality, and processability, which has great design challenges and processing difficulties. As a new micro-nano surface technology, metasurface has shown great potential and the ability to overcome the physical limitations of traditional optical lenses~\cite{Chen2020FlatOW,Engelberg2020TheAO}.
The metasurface is a kind of two-dimensional metamaterial~\cite{padilla2022imaging}. Metamaterials are generally composed of subwavelength metal or dielectric units, which show electromagnetic characteristics different from existing materials in nature, such as negative refraction, optical stealth, and so on~\cite{Liu2011MetamaterialsAN,Smith2000CompositeMW,Shelby2001ExperimentalVO}.
Traditional optical lenses accumulate optical path through the change of thickness and produce phase gradient, to realize the regulation of the wavefront.
When light hits a subwavelength scatterer, its phase will undergo a sudden change, that is, a discontinuous change. By arranging this scatterer into a surface, and then precisely controlling the structure of each unit to control the phase of the light, it is possible to make the light converges to a point~\cite{Yu2011LightPW,Yu2014FlatOW}.
This is called a metalens.
Compared with the traditional optical lens, the metalens has ultra-thin size and weight.
The beam can be focused to the diffraction limit and has an ultra-short focal length~\cite{Khorasaninejad2016MetalensesAV,Lin2020DiffractionlimitedIW}. It has broad application prospects in the visible light band, terahertz band, microwave band, and so on~\cite{Chen2012DualpolarityPM,Chen2016ARO,QuevedoTeruel2019RoadmapOM,Zhang2020LowlossMO,peixia2021single}.

To realize large FoV~\cite{Arbabi2016MiniatureOP,Shalaginov2019ASP,Balli2020AHA,Shalaginov2020SingleElementDF,Fan2020UltrawideangleAH} (Fig.~\ref{fig_10}(c)), achromatic~\cite{Chen2018ABA,Wang2018ABA,Chen2019SpectralTI,Lin2019AchromaticMA}, broadband~\cite{Khorasaninejad2016MultispectralCI,Yang2021WideFA}, high spectral resolution~\cite{Hua2022UltracompactSS} and other characteristics of metalenses, researchers have carried out a series of in-depth studies on metalenses, and several representative works are shown in Table~\ref{tab:Metalens comparison}~\cite{Li2019MetalensBasedMO}.
As an ultrathin optical components~\cite{Tseng2021NeuralNF}, metalens technology provides a new design idea for the panoramic system, and the commercial panoramic imaging system with small volume and high performance will become possible.
  
Metalens has great potential and value in miniaturized imaging, but there are still some challenges~\cite{Wang2021HighefficiencyBA,Pan2022DielectricMF}. In terms of design, the chromatic aberration of the metalens needs to be solved, so the metalens with broad spectrum would still a research hotspot in the future. The focusing efficiency of metalens still needs to be improved~\cite{Patoux2021ChallengesIN}.  In the overall optimization of meta-based imaging systems, it is still difficult to achieve efficient loop data transfer and iteration without the help of custom codes.

In terms of fabrication, the diameter size of metalens needs to be increased in order to integrate into conventional optical systems to realize novel optical systems. In the future, panoramic imaging instruments with hyper-hemispheric FoV can be realized by combining conventional optics and metalens technology to achieve higher-performance miniature panoramic scene perception.

\renewcommand{\arraystretch}{1.5} 
\begin{table*}[tp]  
	
	\centering  
	\fontsize{7.4}{7}\selectfont  
	\begin{threeparttable}  
		\caption{Comparison of different wide FoV metalenses.}  
		\label{tab:Metalens comparison}  
		\begin{tabular}{ccccccc}  
			\toprule 
				Reference (Year) &Focusing efficiency (\%)&Wavelength (nm)&Numerical aperture&Number of metasurface layers&Diffraction-limited FoV ($^\circ$)\\ 

			\midrule  
			Arbabi \textit{et al.}(2016)~\cite{Arbabi2016MiniatureOP}&45-70&850&0.49&2&56\cr
			Groever \textit{et al.}(2017)~\cite{Groever2017MetaLensDI}&30-50&470-660&0.44&2&50\cr  
			Jang \textit{et al.}(2018)~\cite{Jang2018WavefrontSW}&N/A&532&$>$0.5&1&8 mm FoV\cr  	
			Guo \textit{et al.}(2018)~\cite{Guo2018HighEfficiencyAW}&93&Far-field power&0.89&2&120\cr
			Xu \textit{et al.}(2018)~\cite{Xu2018WideangularrangeAH}&95&532&N/A&1&160\cr
			Engelberg \textit{et al.}(2019)~\cite{Engelberg2019NearIRWH}&6-20&825&0.2&1&30\cr
			Shalaginov \textit{et al.}(2020)~\cite{Shalaginov2020SingleElementDF}&32-45&5200&0.24&1&$>$170\cr
			Shalaginov \textit{et al.}(2020)~\cite{Shalaginov2020SingleElementDF}&41-88&940&0.2&1&$\approx$180\cr  
			\bottomrule  
		\end{tabular}  
	\end{threeparttable}  
\end{table*}  

\section{Applications of Panoramic Imaging Systems}
Thanks to the ultra-wide $360^\circ$ FoV offered by panoramic cameras, they have been applied in many scene understanding tasks~\cite{sun2019multimodal}.
In the following subsections, we review the applications with panoramic image systems, for robotics applications like robot navigation and autonomous driving.
\subsection{Semantic Scene Understanding with Image Segmentation}
\begin{figure*}[!t]
\centering
\subfloat{\includegraphics[width=\linewidth]{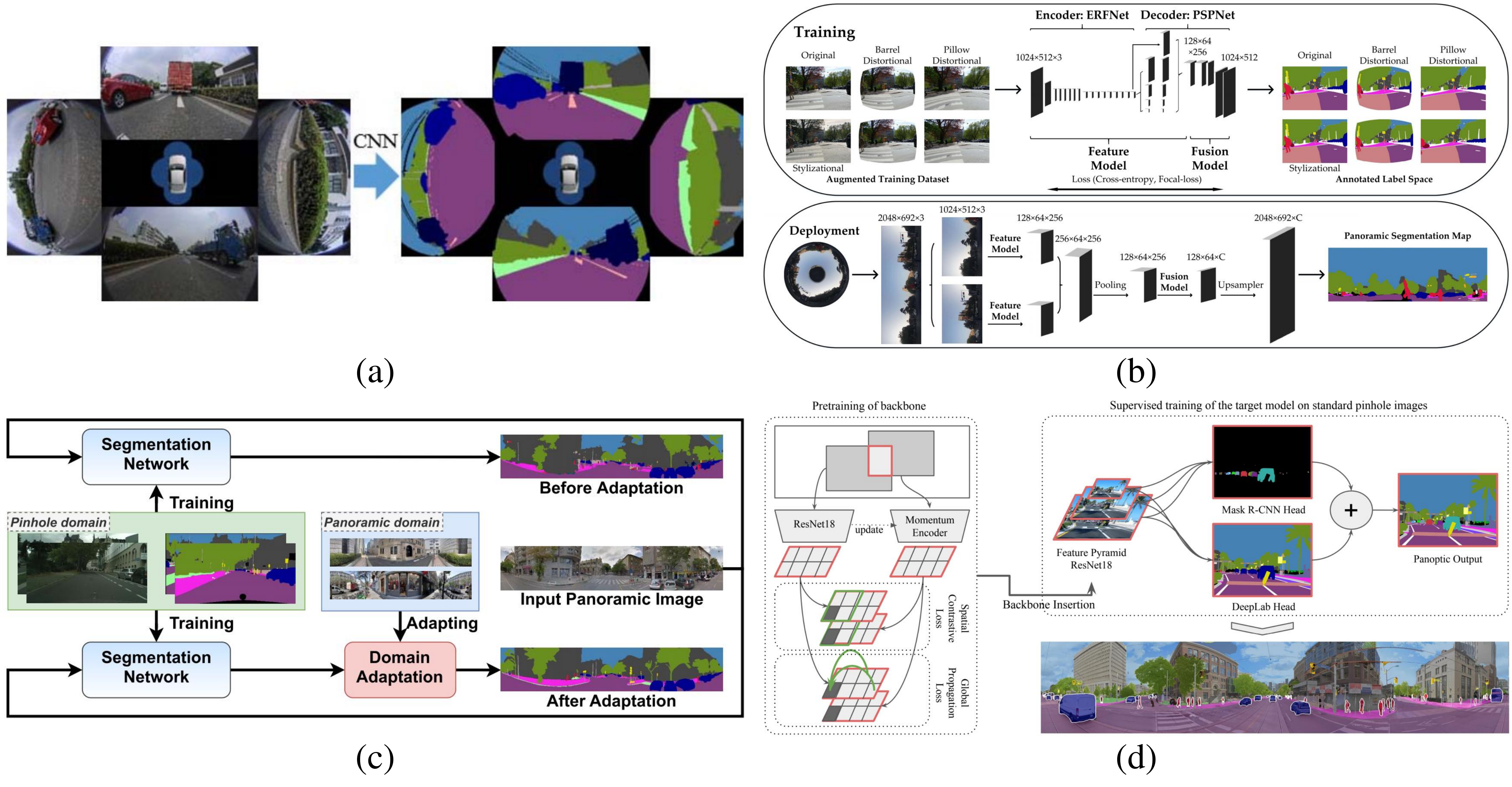}}
\caption{Panoramic scene segmentation for semantic surrounding understanding. (a) Surround-view sensing using multiple fisheye cameras. Reproduced with permission from~\cite{deng2017cnn,deng2019restricted}. Copyright 2017 and 2020, IEEE; (b) Panoramic annular semantic segmentation for entire seamless surrounding perception with a single panoramic camera. Reproduced with permission from ~\cite{yang2019can_we_pass,yang2019pass}. Copyright 2019 and 2020, IEEE; (c) Panoramic semantic segmentation via unsupervised domain adaptation from pinhole images. Reproduced with permission from~\cite{ma2021densepass,zhang2021transfer}. Copyright 2021, IEEE; (d) Panoramic panoptic segmentation for holistic surrounding understanding with dense contrastive pretraining, which provides both pixel-level semantic- and instance information. Reproduced with permission from~\cite{jaus2021panoramic_panoptic}. Copyright 2021, IEEE.}
\label{fig_semantic}
\end{figure*}

Semantic scene perception is an essential task in robotics, as it enables a dense understanding at the pixel level, where the location and category information can be both precisely extracted, offering abundant cues for upper-level navigation operations.
Semantic scene understanding is usually achieved via semantic image segmentation.
When working with $360^\circ$ panoramic cameras, a holistic and comprehensive scene segmentation can be attained for an entire surrounding perception~\cite{yang2019can_we_pass}, as shown in Fig.~\ref{fig_semantic}.

\noindent\textbf{Fisheye camera based semantic segmentation.}
Thanks to the breakthrough of Fully Convolutional Networks (FCNs)~\cite{long2015fcn}, semantic segmentation can be performed in an end-to-end fashion and the field has been significantly advanced with diverse segmentation model architecture- and dataset developments. 
In the wide-FoV semantic segmentation, fisheye- and panoramic cameras have been applied.
Yet, current deep segmentation models are data-driven, and most existing semantic segmentation datasets are established for pinhole, narrow-FoV, distortion-free images.
To train an accurate and robust semantic segmentation model that can generalize well in real-world scenes, large-scale annotated datasets are important.
Annotating semantic labels at the pixel level is extremely labor-intensive and time-consuming, in particular for panoramic images as they have larger distortions, higher complexities due to the wide-FoV, and more small objects in general.
Thereby, most wide-FoV semantic segmentation research works have to deal with the data scarcity issue, aiming to produce robust real-world surrounding parsing models, which are vital for reliable scene understanding.
Deng~\textit{et al.}~\cite{deng2017cnn} first presented a fisheye image semantic segmentation framework (Fig.~\ref{fig_semantic}(a)), which transforms an existing pinhole urban scene segmentation dataset to a synthetic dataset for training and designs an overlapping pyramid pooling based segmentation module to explore both local, global, and region context information.
In particular, the transformation is based on a zoom augmentation method by changing the focal length of the simulated fisheye camera.
They further built a surround-view surrounding road scene segmentation system by using four fisheye cameras and a deformable-convolution-based multi-task learning regimen~\cite{deng2019restricted}, involving real-world and transformed surround-view images for training.
Extending the zoom transformation method, Ye~\textit{et al.}~\cite{ye2020universal} designed an online seven-DoF augmentation method, to synthesize fisheye images captured via sensors of different shooting orientations, postures, and focal lengths, improving the robustness against different distorted fisheye image data.

\noindent\textbf{Panoramic camera based semantic segmentation.}
For panoramic surrounding perception based on a single camera, Yang~\textit{et al.}~\cite{yang2019pass} designed a Panoramic Annular Semantic Segmentation (PASS) framework (Fig.~\ref{fig_semantic}(b)) by using a panoramic annular camera with an entire FoV of $360^\circ{\times}(30^\circ{\sim}105^\circ)$. 
Considering that the distortion in such cameras are lowered in less than $1\%$~\cite{zhou2016comparison_two_panoramic_front} and the imaging model complies with a clear $f$-$\theta$ law, the PASS framework reused large-scale pinhole data for training with a network adaptation method that exploits the correspondence between convolutional features learned from pinhole data and features inferred from panorama segments with a similar FoV.
They achieved seamless semantic segmentation results by using ring-padding and cross-segment padding, together with a variety of data augmentation methods including texture-, geometric-, distortion-, and style augmentation to robustify panoramic segmentation.
DS-PASS~\cite{yang2020dspass} improves PASS by using a more efficient segmentation model and augmenting the detail sensitivity via attention connections between detail-preserved low-level shallow encoder layers and semantic-rich high-level deep decoder layers.
They have deployed the framework on both mobile robotics and instrumented vehicles with a lightweight panoramic annular camera with a FoV of $360^\circ{\times}(50^\circ{\sim}120^\circ)$, and verified the benefits of surrounding semantic sensing for upper-level navigation applications like visual odometry.
Further, multi-source omni-supervised learning was leveraged in~\cite{yang2020omnisupervised} to cover panoramic data in the training stage.
This helps to reach robust segmentation performance, meanwhile bypassing complex panorama separation and network adaptation used in PASS~\cite{yang2019pass} which incurs high computation overhead and cannot make use of the important global context cues as the panorama is divided into discrete segments.
The omni-supervised learning framework was extended by learning spatial positional priors~\cite{yang2021context} and omni-range contextual dependencies~\cite{yang2021capturing} that can stretch across the entire $360^\circ$.
Panoramic semantic segmentation has also been applied with unmanned aerial vehicles in~\cite{sun2021aerial,wang2022high_performance} for drone-perspective panoramic remote sensing with a FoV of $360^\circ{\times}(30^\circ{\sim}100^\circ)$.

\noindent\textbf{Panoramic segmentation from an adaptation perspective.}
More recently, to address the dearth of annotated panoramic images, researchers have explicitly formalized panoramic segmentation as an unsupervised domain adaptation problem or a domain generalization problem by transferring from the data-rich pinhole domain to the data-scarce panoramic domain~\cite{ma2021densepass,jaus2021panoramic_panoptic}. 
For domain adaptation, one can use labeled pinhole data as the source domain and unlabeled panoramic images as the target domain.
For domain generalization, one can only use the source domain images, with the aim to produce a robust, generalized segmentation model in the target domain. 
P2PDA~\cite{ma2021densepass,zhang2021transfer} (Fig.~\ref{fig_semantic}(c)) designed attention-augmented domain adaptation modules to detect and magnify the pinhole-panoramic correspondences in multiple spaces.
Jaus~\textit{et al.}~\cite{jaus2021panoramic_panoptic,jaus2022panoramic_panoptic_segmentation_insights} introduced panoramic panoptic segmentation, which extended panoramic semantic segmentation by also offering instance-level understanding (Fig.~\ref{fig_semantic}(d)).
It outputs both pixel-level semantic classes and instance IDs, and thereby enabling a more holistic surrounding perception.
By appending a dense contrastive pretraining stage, they obtained significant generalization gains when segmenting unseen panoramic images.
Panoramic panoptic video segmentation is also addressed with the Waymo open dataset~\cite{mei2022waymo} covering a FoV of $220^\circ$.
Distortions in panoramic images have been taken into considerations. Many researchers opt to directly perform segmentation models on the equirectangular panoramic images with distortions and come up with distortion-aware model designs.
Hu~\textit{et al.}~\cite{hu2022distortion} considered the panoramic image generation process and tackled distortions via a deformable-convolution-based module.
Zhang~\textit{et al.}~\cite{zhang2022bending,zhang2022behind} developed a distortion-aware transformer model for adapting to panoramic images. The transformer model is integrated with both deformable patch embedding and deformable MLP modules, which well tackle the distortions and enhance the robustness of both indoor and outdoor panoramic segmentation.
Zheng~\textit{et al.}~\cite{zheng2022complementary_bidirectional} used complementary feature representations to provide implicit distortion distribution priors for guiding indoor panoramic semantic segmentation.
Moreover, there are many other fisheye~\cite{yogamani2019woodscape,kumar2021omnidet,eising2021near_field_perception,ahmad2022fisheyehdk}, panoramic~\cite{xu2019semantic_synthetic,orhan2021semantic_segmentation_outdoor,liu2021pano_sfmlearner}, and omnidirectional~\cite{zhang2019orientation_aware,sekkat2020omniscape,sekkat2022comparative_semantic_omnidirectional,li2022omnicity} semantic segmentation works in the field, evidencing the relevance of wide-FoV semantic scene understanding for real-world applications.

As panoramic images usually have a large FoV and high resolution, researchers have also taken this into account by designing compact models to enable fast response for efficiency-critical applications. For example, PASS~\cite{yang2019pass} and DS-PASS~\cite{yang2020dspass} use efficient backbones, but enhance the performance by using attention blocks either in the decoder or in the lateral connections.
The framework in~\cite{yang2020omnisupervised} makes use of knowledge distillation to boost the performance of the compact models.
The panoramic panoptic segmentation framework~\cite{jaus2022panoramic_panoptic_segmentation_insights} also substitutes all investigated panoptic backbones with the efficient ResNet-18~\cite{he2016deep} to support mobile agents operating in real-world traffic scenarios. They also show that their efficient modifications become more important when feeding images of higher resolutions into the model.

\subsection{Geometric Scene Understanding with Depth Estimation}
\begin{figure*}[!t]
\centering
\subfloat{\includegraphics[width=\linewidth]{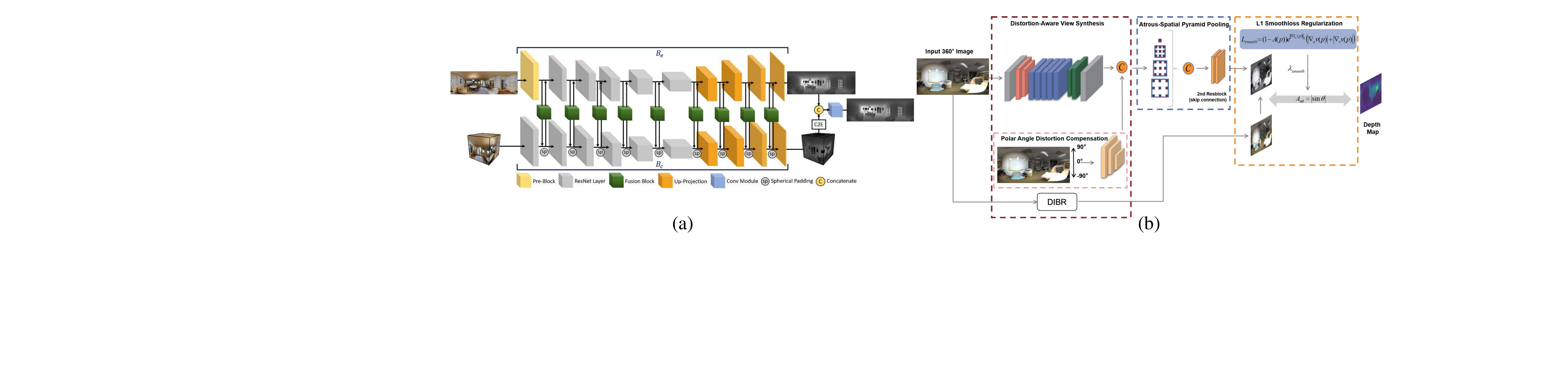}}
\caption{Panoramic depth estimation for geometric surrounding understanding. (a) Example of a supervised method, which bidirectionally fuses representations of equirectangular- and cube-map projections for monocular $360^\circ$ depth estimation. Reproduced with permission from~\cite{wang2020bifuse}. Copyright 2020, IEEE; 
(b) Example of an unsupervised method, which leverages a combination of an auto-encoder based on convolutional neural networks, Depth-Image-Based Rendering (DIBR), a polar angle distortion compensation layer, and atrous-spatial pyramid pooling. Reproduced with permission from~\cite{lai2021olanet}. Copyright 2021, IEEE.}
\label{fig_depth}
\end{figure*}

Panoramic depth estimation is another important task, which aims to predict dense depth maps based on a panoramic camera, and it is relevant for many robotics applications as it enables computer systems to obtain $360^\circ$ geometric information and understand the surrounding 3D world. 
There are two main categories of panoramic depth estimation methods, which are supervised and unsupervised methods, whose representative examples are provided in Fig.~\ref{fig_depth}.

\noindent\textbf{Supervised methods.}
``Supervised'' means that the depth estimation methods directly have the corresponding ground-truth depth information to learn. 
In the supervised category, OmniDepth~\cite{zioulis2018omnidepth} generated synthetic panoramic datasets for supervised learning, via rending existing large-scale 3D datasets, which enables direct depth estimation on omnidirectional content in the form of equirectangular images.
DistConv~\cite{tateno2018distortion} put forward distortion-aware convolution filters, which endows a pinhole-trained model to smoothly work on equirectangular projection of panoramic data.
In~\cite{de2018eliminating,zhou2020padenet}, projection transformation is used to adapt existing rectilinear image datasets into equirectangular ones for learning panoramic depth estimation.
They also leverage ring padding to enhance the continuity at the panorama border for a seamless estimation, which is similarly considered in panoramic semantic segmentation~\cite{yang2019can_we_pass,yang2019pass}
In~\cite{jin2020geometric_structure_depth_estimation}, geometric structures are used as a prior to aid supervised panoramic depth estimation.
Moreover, HoHoNet~\cite{sun2021hohonet} and SliceNet~\cite{pintore2021slicenet} first squeezed the extracted feature maps into a horizontal 1D representation due to the assumption that indoor panoramas are aligned to the gravity vector, and then recovered the dense depth map predictions in the equirectangular projection.
Unlike previous methods that calculated the shapes of convolution kernels according to the latitude coordinates~\cite{tateno2018distortion} in a fixed manner, ACDNet~\cite{zhuang2021acdnet} combined the convolution kernels with different dilation rates in an adaptive manner and improved the estimation in the equirectangular projection.
There are also fusion-driven methods with the aim to exploit the advantages of both equirectangular- and cube-map projections~\cite{wang2020bifuse,jiang2021unifuse,bai2022glpanodepth}, with methods including bidirectional fusion~\cite{wang2020bifuse,wang2022bifuse++}, unidirectional fusion~\cite{jiang2021unifuse}, and usage of an asymmetric architecture that leverages a transformer model in the cube-map branch and a convolutional neural network in the equirectangular branch~\cite{bai2022glpanodepth}.
OmniFusion~\cite{li2022omnifusion} transforms a $360^\circ$ panorama into perspective image patches (\textit{i.e.}, tangent images) to attain patch-wise predictions and fuses the image features with 3D geometric features to enhance depth estimation quality. Such tangent image representations have also been used in~\cite{rey2022360monodepth} for estimating high-resolution depth maps.
Regarding the advantages of the projection representations, equirectangular projection provides full-FoV observations and cube-map projection offers distortion-free representations~\cite{wang2020bifuse} (see Fig.~\ref{fig_depth}(a)).
More recently, PanoFormer~\cite{shen2022panoformer} put forward a panoramic transformer model to perform depth estimation, where the patch division is conducted in the spherical tangent domain to tackle distortions and geometric structures are modeled by using adding learnable token flows in the self-attention layers.

\noindent\textbf{Unsupervised methods.}
``Unsupervised'' means that the depth estimation methods do not require ground-truth depth information at the training time and thereby reduce the data preparation burden, which are beneficial for real applications.
Unsupervised panoramic depth estimation is less visited compared to the supervised category, but it started to receive more attention.
Zioulis~\textit{et al.}~\cite{zioulis2019spherical} formalized a full spherical disparity model and established a new untouched supervision scheme for spherical view synthesis using depth-image-based-rendering combined with spherical attention.
Thereby, they facilitated a self-supervised monocular $360^\circ$ depth learning paradigm via spherical view synthesis. 
OlaNet~\cite{lai2021olanet} made an early self-supervised attempt to predict depth information in the equirectangular projection of $360^\circ$ images.
They leveraged a combination of an anto-encoder based on convolutional neural networks, distortion-aware spherical view synthesis with depth-image-based rendering, and polar angle distortion compensation, to tackle the challenging self-supervised $360^\circ$ depth prediction task (Fig.~\ref{fig_depth}(b)).
Zhou~\textit{et al.}~\cite{zhou2021panoramic} extended PADENet~\cite{zhou2020padenet} and combined unsupervised and supervised learning to improve panoramic depth estimation specifically in indoor scenes.
In~\cite{yun2021improving_joint}, gravity-aligned videos were used to facilitate self-supervised $360^\circ$ depth learning.
They also combined supervised- and self-supervised learning to compensate the weaknesses of each learning method towards more accurate depth estimation.
More recently, in~\cite{yan2022multimodal_masked_panoramic_depth_completion}, multi-modal masked pretraining was used to enhance panoramic depth completion, which aims to complete the depth channel of a single $360^\circ$ RGB-D pair captured from a panoramic 3D camera.
Panoramic depth estimation and completion were also potentially useful for improving the perception ability and operational efficiency of mobile robots in large-scale environments~\cite{liu2022cross_modal_depth_estimation}.

\subsection{Visual Localization, Odometry, and SLAM}
\begin{figure*}[!t]
\centering
\subfloat{\includegraphics[width=\linewidth]{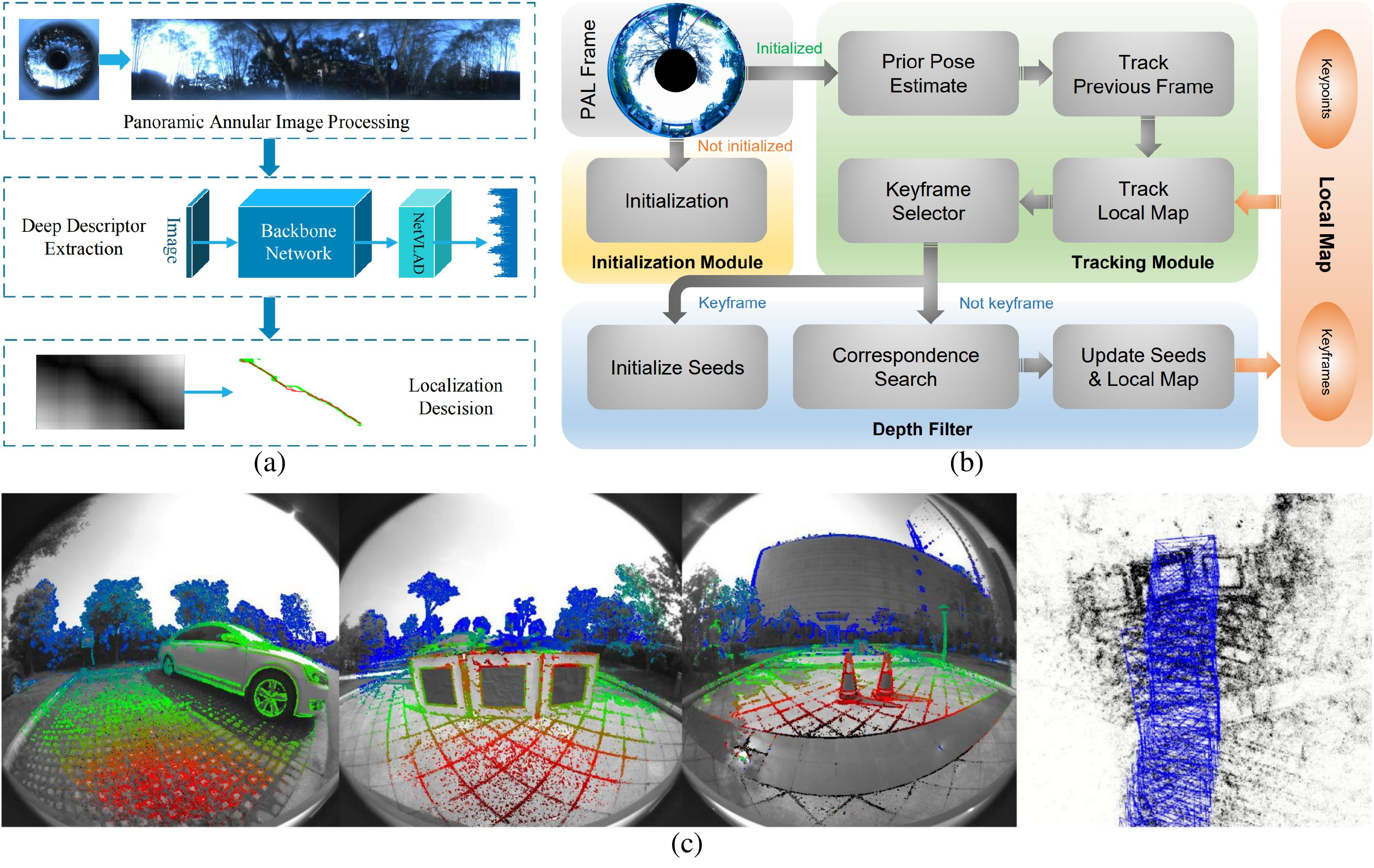}}
\caption{Panoramic visual localization, visual odometry, and visual SLAM. (a) Example of a panoramic visual localization method, where panoramic mages are processed and fed into a NetVLAD network~\cite{arandjelovic2016netvlad} to form the active deep descriptor and sequential matching is utilized to generate the localization result. Reproduced with permission from~\cite{cheng2019panoramic}. Copyright 2016 and 2019, IEEE; (b) Example of a panoramic visual odometry with specifically designed initialization, tracking, and depth filter modules, for tackling rapid motion and dynamic scenarios. Adapted with permission from~\cite{chen2019palvo}. Copyright 2019, Optical Society of America; (c) Example of a panoramic visual SLAM method, where the captured omnidirectional RGB images are overlaid with color-coded semi-dense depth information and the 3D depth map is reconstructed. Reproduced with permission from~\cite{wang2018realtime_omnidirectional_slam}. Copyright 2018, IEEE.}
\label{fig_vo}
\end{figure*}

In this subsection, we review important applications of panoramic cameras related to the localization and mapping in robotic environments, including panoramic visual localization, panoramic visual odometry, and panoramic Simultaneous Localization And Mapping (SLAM), as shown in Fig.~\ref{fig_vo}.

\noindent\textbf{Panoramic visual localization.}
Visual localization is a task that recovers the camera location from database images based on the query images, which is important for many applications like intelligent vehicles and assisted pedestrian navigation~\cite{cheng2019panoramic}.
To overcome the challenges in matching query and database images with various appearance variations like illumination variations, moving object variations, and orientation variations, panoramic cameras are usually applied.
In~\cite{iscen2017panorama}, location recognition was revisited by conducting a panorama-to-panorama matching process and they found that it is preferable to aggregate individual pinhole views into a vector for matching, rather than extracting a single descriptor from an explicit panorama.
In~\cite{siva2018omnidirectional}, omnidirectional observation was integrated to perform visual localization and found that the front- and back views are the most descriptive for long-term place recognition. 
Cheng~\textit{et al.}~\cite{cheng2019panoramic} proposed a Panoramic Annular Localizer (see Fig.~\ref{fig_vo}(a)), which incorporates panoramic annular lens and active deep image descriptors in a visual localization system.
CFVL~\cite{fang2020cfvl,fang2020panoramic_localizer} designed a coarse-to-fine visual localization framework by using equirectangular representations in the coarse stage to keep feature consistency with the panorama-trained descriptor, and cube-map representations in the fine stage of key-point matching to conform to the planarity of the images.
In~\cite{jayasuriya2020active_perception_localisation}, a localization framework is designed by using high-level semantic information such as detected landmarks and ground surface boundaries, which are obtained via an omnidirectional camera, and their localization is conducted via extended Kalman filter.
In~\cite{orhan2022semantic_pose_verification}, semantic similarity was checked between query- and database images based on contrastive learning on a dataset of semantically-segmented images.
Their database images are generated from panoramas which present wide-FoV that facilitates to correctly localize the query images where standard-FoV cameras may fail with non-overlapping FoVs. 
Moreover, rotation equivariant orientation estimation~\cite{zhang2020rotation_equivariant} and semantic graph embedding lifted in 3D space~\cite{zhang2021lifted_semantic_graph} are explored for panoramic place recognition based on omnidirectional images.

\noindent\textbf{Panoramic visual odometry.}
Visual odometry estimates the egomotion of an agent by using only the continuous images captured by cameras attached to it~\cite{chen2019palvo}.
One of the shared premise in visual odometry methods is that there must be enough overlaps between two adjacent frames for computing the relative motion.
Pinhole-camera-based methods inevitably suffer in rapid-motion scenarios or dynamic scenarios with moving objects which may occupy most of the limited FoV.
Incorporating panoramic observations is potential to help address the challenges in such scenarios.
In~\cite{zhang2016benefit_large_fov}, the benefits of using large-FoV cameras (\textit{e.g.}, fisheye or catadioptric) were investigated. They pointed out that wide-angle observations are advantageous in indoor scenes with more evenly-distributed features and the camera can help track features for long-term autonomy.
However, in outdoor scenes, the loss of angular resolution for wider FoVs is intensified due to the higher depth range, and using large-FoV cameras is less accurate.
PVO~\cite{lin2018pvo} took advantage of panoramic cameras by modifying the initialization process in the visual odometry and redefining the re-projection errors according to the panoramic camera model, which achieved higher precision in translation and rotation.
Further, stereo omnidirectional cameras have been put use in~\cite{matsuki2018omnidirectional_dso,jaramillo2019visual_odometry_stereo_omnidirectional}.
For example, dual fisheye cameras are used in~\cite{matsuki2018omnidirectional_dso} to obtain an ultra-wide FoV that ensures big image overlaps between frames and more spatially distributed points, where their joint optimization model helped to attain improved robustness over a pinhole camera model.
Chen~\textit{et al.}~\cite{chen2019palvo} presented PALVO (Fig.~\ref{fig_vo}(b)) by applying a Panoramic Annular Lens.
Based on the panoramic annular camera model, they adapted initialization-, tracking-, and depth filter modules, which enabled PALVO to reach high robustness to rapid motion and dynamic objects. 
In~\cite{chen2021semantic}, semantic information were used to further enhance PALVO, with semantic-guided keypoint extraction and correspondence search along epipolar curve, as well as semantic-based examination in pose optimization, which achieved more robust pose estimation.
The Direct Sparse Odometry (DSO)~\cite{engel2017direct} has also been augmented by attaining omnidirectional perception towards real-world applications~\cite{matsuki2018omnidirectional_dso,huang2022360vo,hu2019indoor_positioning}.
For example,
panoramic visual odometry has been applied with wearable devices to help people with visual impairments in indoor positioning and navigation~\cite{hu2019indoor_positioning}.

\noindent\textbf{Panoramic visual SLAM.}
Visual SLAM is a technology that estimates the poses of cameras and reconstructs the scenes, which usually extends the methods of visual odoemtry.
In~\cite{caruso2015large_direct_slam}, a large-scale direct SLAM method is developed for panoramic- or wide-FoV fisheye cameras by directly formulating the tracking and mapping according to the omnidirectional camera model.
By using two different omnidirectional imaging models, their system makes it possible to use a broad range of classical dioptric- or catadioptric imaging systems. 
They have verified their approach on data captured by a $185^\circ$-FoV fisheye lens, and observed improved localization accuracy and enhanced robustness against fast camera rotations compared to the baseline approach using a standard camera.
In~\cite{wang2018realtime_omnidirectional_slam}, the ORB-SLAM2 framework~\cite{mur2017orb_slam2} was extended by utilizing a unified spherical camera model and the semi-dense feature mapping.
They validated their approach on real-world data collected by fisheye cameras set up on a car, of which the capturing FoV is around $185^\circ$ (see Fig.~\ref{fig_vo}(c)).
Panoramic annular camera has been applied in~\cite{chen2021panoramic,wang2022pal_slam}. 
PA-SLAM~\cite{chen2021panoramic} extended PALVO~\cite{chen2019palvo} to a full panoramic visual SLAM system with loop closure and global optimization. They used a panoramic annular lens with a FoV of $360^\circ{\times}(30^\circ{\sim}92^{\circ})$ and a global shutter camera for experiments. They confirmed the better performance of PA-SLAM with respect to ORB-SLAM2~\cite{mur2017orb_slam2} and a cube-map-based fisheye SLAM method~\cite{wang2018cubemapslam} on real-world data.
PAL-SLAM~\cite{wang2022pal_slam} designed a feature-based SLAM for panoramic annular lens with a FoV of $360^\circ{\times}(45^\circ{\sim}85^{\circ})$.
They used a filtering mask to acquire features from the annular useful area, in order to conduct carefully-designed initialization, tracking, mapping, loop detecting, and global optimization. 
In this pipeline, they utilized the calibrated panoramic annular camera model to translate the mapped features onto a unit vector for the subsequent algorithms. 
The robust localization performance of PAL-SLAM are confirmed on large-scale outdoor and indoor panorama sequences, showing the advantages in comparison to a traditional SLAM method based on a camera with a narrow FoV.
Additionally, there are some more mapping systems like landmark-tree mapping~\cite{jager2013efficient_navigation} and freespace mapping~\cite{lukierski2015rapid_freespace_mapping_omnidirectional,posada2018semantic_mapping_omnidirectional_vision} based on a single omnidirectional camera,
360ST-Mapping~\cite{liu360st_mapping} based on online semantic-guided scene topological map reconstruction,
as well as UPSLAM~\cite{cowley2021upslam} based on a graph of panoramic images.

Furthermore, panoramic visual odometry and SLAM have also been implemented based on multiple sensors, \textit{e.g.}, using multi-cameras~\cite{seok2019rovo,seok2020rovins,won2020omnislam}, with inertial sensors~\cite{ramezani2017omnidirectional_vio_karlman,ramezani2018pose_omnidirectional_vio,gao2021lovins,wang2022lf_vio}, GPS~\cite{yu2019gps}, and LiDAR sensors~\cite{kang2021rpv_slam}.
Among them, LF-VIO~\cite{wang2022lf_vio,wang2022lfvio_loop} put forward a universal framework that can work with diverse large-FoV cameras, where imaging points may even appear on the negative half plane. They verified the effectiveness of their visual-inertial-odometry framework on panoramic annular cameras with a FoV of $360^\circ{\times}(40^\circ{\sim}120^\circ)$ and fisheye cameras with a FoV of $360^\circ{\times}(0^\circ{\sim}93.5^\circ)$, as well as sensor fusion systems with LiDAR measurements.

\subsection{Optical Flow Estimation, Object Detection, and Others}
\begin{figure*}[!t]
\centering
\subfloat{\includegraphics[width=\linewidth]{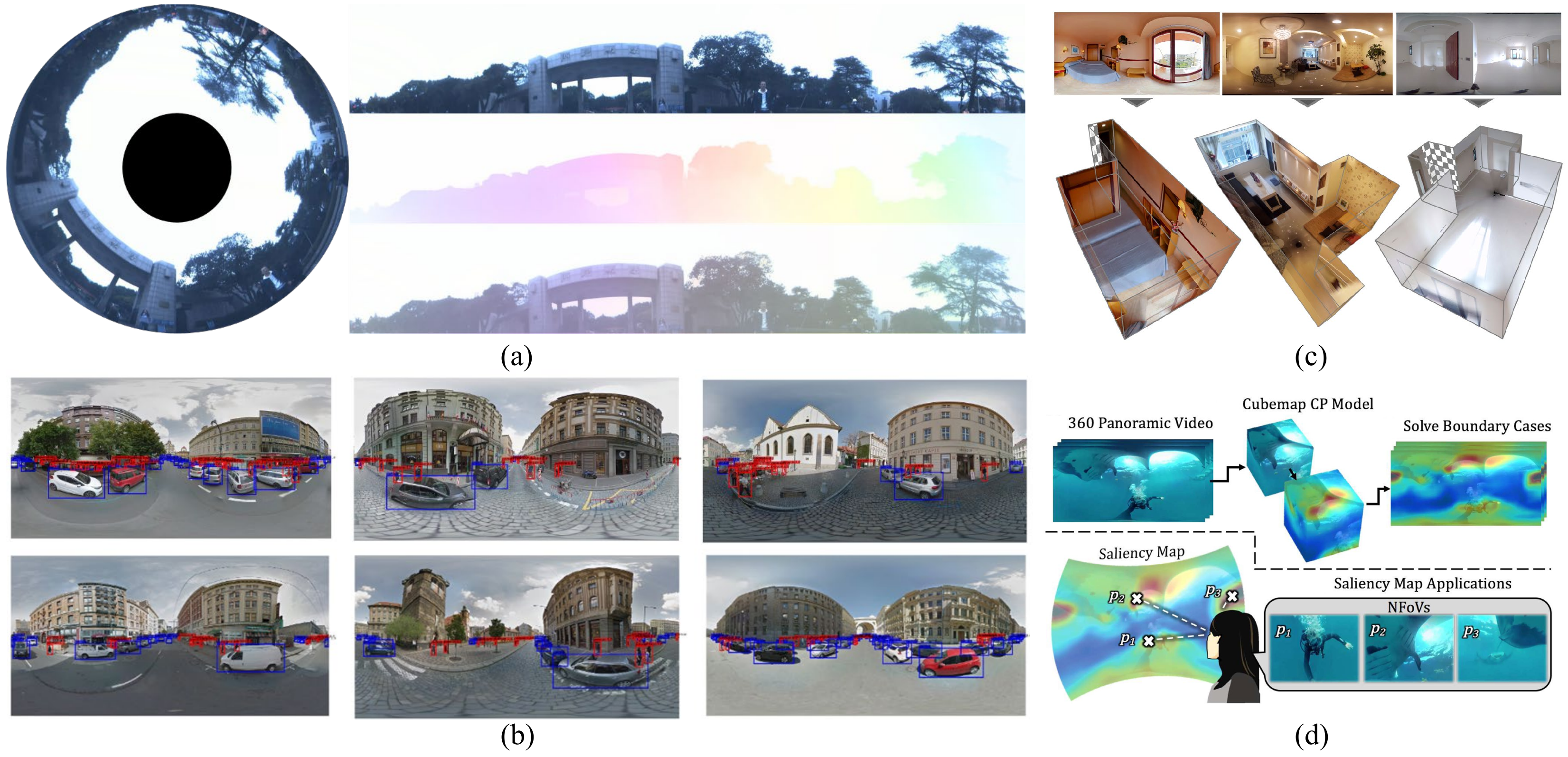}}
\caption{Panoramic flow estimation, object detection, layout estimation, and saliency prediction. (a) Example of panoramic flow estimation on real-world surrounding view for $360^\circ$ scene temporal understanding. Reproduced with permission from Shi~\textit{et al.}~\cite{shi2022panoflow}. (b) Example of panoramic object detection to identify objects and locations of interest in surroundings. Reproduced with permission from~\cite{wang2019object_detection_curved}. Copyright 2019, IEEE; (c) Example of panoramic 3D layout estimation for indoor scene understanding. Reproduced with permission from~\cite{yang2019dula}. Copyright 2019, IEEE; (d) Example of panoramic saliency prediction to reflect the degree of importance of a pixel in $360^\circ$ contents to the human visual system. Reproduced with permission from~\cite{cheng2018cube}. Copyright 2018, IEEE.}
\label{figure_flow}
\end{figure*}

In addition to semantic segmentation and depth estimation, there are many potential information worth mining in panoramic images, such as optical flow estimation and object detection, as shown in Fig.~\ref{figure_flow}. In this subsection, we review these panoramic vision tasks respectively.

\noindent\textbf{Optical flow estimation.}
Optical flow reflects pixel-by-pixel correspondence on time-series images and can be obtained by learning-based methods~\cite{dosovitskiy2015flownet}. Flow estimation in panoramic content is an essential task, which enables autonomous vehicles and robots to acquire temporal cues of the surrounding scenes.
Due to the lack of large-scale panoramic optical flow datasets, research on $360^\circ$ flow estimation mainly focuses on the domain adaptation.
In order to leverage existing perspective models and benefit from advances in optical flow networks, two common perspectives are adapting models and datasets respectively.
OmniFlowNet~\cite{artizzu2021omniflownet} adapted the famous LiteFlowNet~\cite{hui2018liteflownet} that designed for pinhole cameras to panoramic images and does not require retraining.
Considering the equirectangular distortion imported by the spherical camera, they used the distortion-aware convolution to replace the regular grid-based 2D convolution via revising the coordinates of the convolution kernel.
To address the distortion that introduced by spherical-to-plane projection in panoramic content,
Bhandari~\textit{et al.}~\cite{bhandari2021revisiting} proposed a spherical data augmentation to project perspective flow field onto the sphere and then reprojected to the equirectangular format. 
However, this data augmentation is offline and not flexible, and the perspective dataset should be transformed in advance for training.
Yuan~\textit{et al.}~\cite{yuan2021360} introduced a $360^\circ$ flow optical flow method by estimating on the tangent images.
They directly use the off-the-shelf perspective optical flow method to estimate the flow field of panoramic images, and then apply the optical flow estimated under cubemap and icosahedron projection to benefit the panoramic flow estimation.
OmniFlow~\cite{seidel2021omniflow} explored the human omnidirectional optical flow in a 3D indoor environment with a rendering engine, and generated a synthetic $360^\circ$ human optical flow dataset.
Among these approaches, PanoFlow~\cite{shi2022panoflow} (see Fig.~\ref{figure_flow}(a)) presented a generic cyclic flow estimation framework, which can be adapted to any existing perspective learning-based model, \textit{e.g.,} RAFT~\cite{teed2020raft} and CSFlow~\cite{shi2022csflow}.
They generated the publicly available panoramic flow dataset Flow360 under city street scenes to facilitate training and quantitative analysis, and verified the precision and efficiency of their approach via a panoramic annular camera with a FoV of $360^\circ{\times}(30^\circ{\sim}90^\circ)$.

\noindent\textbf{Object detection.}
Object detection in the surrounding environment is also a demanded task as the increasing interest in autonomous mobile systems such as drones and assistant robots. Early panoramic object detection systems\cite{cinaroglu2016direct_object_detection,wang2009object} used hand-designed operators, and the subsequent emergence of CNN has given new vitality to this field. 
Garanderie~\textit{et al.}~\cite{de2018eliminating} extended the Faster R-CNN~\cite{ren2015faster} to support 3D object pose regression.
Without changes to the network architecture, their extended model can be used on either rectilinear or equirectangular content by recovering 3D position using image transformation matrix and spherical coordinate projection.
Yang~\textit{et al.}~\cite{yang2018object_equirectangular} leveraged VR videos on YouTube and proposed a bounding-FoV annotation to create a equirectangular panorama dataset for object detection.
They then adapted YOLO~\cite{redmon2017yolo9000} to panoramic contents with re-projections to realize panoramic object detection.
Wang~\textit{et al.}~\cite{wang2019object_detection_curved} (Fig.~\ref{figure_flow}(b)) proposed a distortion data augmentation method that uses the cropped car images to randomly back project in the equirectangular image plane.
Considering that distortion of large objects is more pronounced, they aggregated the position information of anchor box into the network.
Guerrero-Viu~\textit{et al.}~\cite{guerrero2020whats_in_my_room} took advantage of the off-the-shelf layout estimation method~\cite{fernandez2020corners} to simultaneously obtain the 2D object segmentation masks and the 3D bounding boxes. They also replaced all standard convolutions with EquiConvs~\cite{fernandez2020corners} to implicitly adapt convolution kernels' shape and size according to the distortions. 

\noindent\textbf{Other downstream tasks.} In addition to the above applications, panoramic cameras are also integrated with deep learning algorithms in other fields.
\emph{Layout Estimation}: The 3D layout of a room specifies the positions, orientations, and the heights of the walls according to the vision point.
Estimating the layout of a room by a single panorama can benefit robotics and VR/AR applications.
LayoutNet~\cite{zou2018layoutnet} is a deep CNN that estimates the layout of an indoor scene from a single perspective or a panoramic image.
They extended the annotations for the Stanford2D3D dataset~\cite{armeni2017joint} with room layout annotations.
DuLa-Net~\cite{yang2019dula} (Fig.~\ref{figure_flow}(c)) fused the estimation under equirectangular images and perspective ceiling images via a two-stream network architecture.
They also introduced Realtor360 dataset, which consist of Manhattan-world room layouts in panoramic format.
CFL~\cite{fernandez2020corners} proposed EquiConvs by deforming the shape of the kernels according to the geometrical priors of panoramas.
It can predict complex geometries like non-Manhattan structures.
More recently, DeepPanoContext\cite{zhang2021deeppanocontext} was introduced as the first learning-based pipeline that recovers 3D room layout and detailed shape, pose, and location for objects in the scene from a single panoramic image for holistic 3D scene understanding.
\emph{Anomalous Detection}: Anomalous detection is a task to detect anomalous objects that in a specific scene or behaving dangerously, which is important for intelligent panoramic video surveillance systems.
Benito-Picazo~\textit{et al.}~\cite{benito2021deep_anomalous} developed a low energy consumption system on a raspberry for video anomalous object detection.
The panorama videos are obtained via a Point Grey Ladybug 3 Spherical camera and then fed to the CNN to determine whether it contains any anomalous object.
\emph{Saliency Prediction}: Saliency prediction is a popular computer vision task which means predicting where humans look in an given image~\cite{liu2010learning}. Consecutive saliency prediction in panoramic videos is vital for applications such as viewpoint guidance.
Zhang~\textit{et al.}~\cite{zhang2015exploiting} explored the utility
of the surrounding cue for saliency detection, and proposed BMS, which is a saliency model using simple image processing operations to leverage the topological structural cue in panoramas.
Cheng~\textit{et al.}~\cite{cheng2018cube} (Fig.~\ref{figure_flow}(d)) put forward a weakly-supervised spatial-temporal saliency prediction model to overcome large viewpoint variations in $360^\circ$ videos. They also collected a Wild-360 dataset where one-third frames are annotated with saliency heatmaps.

\section{Conclusions and perspectives}
The panoramic imaging optical system is given a high expectation of future intelligent perception sensor. It has shown great application potential in current scene understanding applications such as autonomous driving and robotics systems.
Associated with the expectations are the challenges of design and fabrication, compressing a high-performance panoramic system into a compact space for multidimensional panoramic information acquisition. Although the parametric limitations of optical systems have constituted a huge obstacle on this path, innovative optical architectures, and emerging optical technologies have lit the light of this path. The rise of ultra-precision manufacturing provides strong support for micro-nano processing such as freeform surfaces, thin-plate optics, and metalens. These new optical engines provide an effective alternative to traditional optical components and support more design freedom to achieve new capabilities and high performance beyond traditional geometric optics. In the future, the panoramic imaging system will combine these emerging technologies towards compact, intelligent and multidimensional development to achieve more powerful panoramic scene environment perception and surrounding understanding.

This review article conducts a systematic and detailed literature analysis on the knowledge of panoramic optical systems and downstream perception applications, and clarifies the advantages, progress, limitations, future challenges, and directions for the improvement of panoramic optical systems. The article identifies the valuable challenges that panoramic perception needs to face from the aspects of optical system and algorithm application, including the following:

1) The new optical surface technology will contribute to the miniaturization, lightweight, and high performance of panoramic imaging systems.

2) Multi-FoV, multi-functional (\textit{e.g.} zoom, polarization), high temporal resolution, and high spectral multidimensional perception of the panoramic optical systems will facilitate more new applications of scene understanding.

3) Based on the fusion of new intelligent sensors and computational imaging, the development of panoramic optical instruments for scene understanding will be promoted.

4) Combining a panoramic optical system with an ultra-wide FoV and an event sensor with an ultra dynamic range or an active sensing LiDAR with a same projection behavior for multi-modal robust surrounding perception.

5) Considering that the annotation on the entire panoramas is difficult and expensive to obtain, exploring algorithms with unsupervised or weakly supervised learning, or adapted from the existing pinhole domain to the panoramic domain will benefit practical applications.

In general, with the unique advantages of a $360^\circ$ large FoV, the panoramic optical systems will combine emerging sensors, ultra-precise optical manufacturing, computational imaging, and artificial intelligence to achieve ultra-high performance, new functions, small volume, and multidimensional sensing and low-power panoramic intelligent instruments to enhance the way humans observe the world and have a deeper and more realistic perception of the surroundings. The panoramic imaging system has transformed from traditional optics to intelligent computing optics, and will be applied in various applications such as autonomous driving, medical health, security monitoring, deep space exploration, deep sea exploration, celestial terrain detection, and intelligent robots, greatly improving our life and the way to perceive the environment.

\bibliographystyle{IEEEtran}
\bibliography{bib}

\end{document}